\documentclass{article}
\usepackage{microtype}
\usepackage{wrapfig}
\usepackage{graphicx}
\usepackage{subcaption}
\usepackage{booktabs}
\usepackage{tabularx}
\usepackage{hyperref}

\usepackage[accepted]{icml2024}
\usepackage{amsmath}
\usepackage{amssymb}
\usepackage{mathtools}
\usepackage{amsthm}
\usepackage[capitalize,noabbrev]{cleveref}

\newcommand{\mbf}[1]{{\boldsymbol{\mathbf{#1}}}}
\newcommand{\bm}{\mbf}

\theoremstyle{plain}

\theoremstyle{definition}

\theoremstyle{remark}

\usepackage[textsize=tiny]{todonotes}
\usepackage{xspace}
\usepackage{physics}

\definecolor{amber}{RGB}{206,18,86}
\definecolor{blueish}{RGB}{43,140,190}
\definecolor{purpleish}{RGB}{140,81,10}
\definecolor{ye}{HTML}{ff7f00}
\definecolor{pu}{HTML}{984ea3}
\definecolor{gre}{HTML}{4daf4a}
\definecolor{re}{HTML}{e41a1c}
\newcommand{\tc}[1]{\textcolor{purple}{#1}}

\newcommand{\mup}{$\mu$P\xspace}
\newcommand{\R}{\mathbb{R}}
\newcommand{\din}{{d_\mathrm{in}}}
\newcommand{\dout}{{d_\mathrm{out}}}

\icmltitlerunning{Compute Better Spent: Replacing Dense Layers with Structured Matrices}

\begin{document}

\twocolumn[

\icmltitle{Compute Better Spent: Replacing Dense Layers with Structured Matrices}

\icmlsetsymbol{equal}{*}

\begin{icmlauthorlist}
\icmlauthor{Shikai Qiu}{equal,nyu}
\icmlauthor{Andres Potapczynski}{equal,nyu}
\icmlauthor{Marc Finzi}{cmu}
\icmlauthor{Micah Goldblum}{nyu}
\icmlauthor{Andrew Gordon Wilson}{nyu}
\end{icmlauthorlist}

\icmlaffiliation{nyu}{New York University}
\icmlaffiliation{cmu}{Carnegie Mellon University}
\icmlcorrespondingauthor{Shikai Qiu}{sq2129@nyu.edu}
\icmlcorrespondingauthor{Andres Potapczynski}{ap6604@nyu.edu}
\icmlcorrespondingauthor{Marc Finzi}{maf820@nyu.edu}
\icmlcorrespondingauthor{Micah Goldblum}{goldblum@nyu.edu}
\icmlcorrespondingauthor{Andrew Gordon Wilson}{andrewgw@cims.nyu.edu}

\icmlkeywords{Machine Learning, ICML}
\vskip 0.3in
]

\printAffiliationsAndNotice{\icmlEqualContribution}

\begin{abstract}
Dense linear layers are the dominant computational bottleneck in foundation models. Identifying more efficient alternatives to dense matrices has enormous potential for building more compute-efficient models, as exemplified by the success of convolutional networks in the image domain.
In this work, we systematically explore structured matrices as replacements for dense matrices. We show that different structures often require drastically different initialization scales and learning rates, which are crucial to performance, especially as models scale. Using insights from the Maximal Update Parameterization, we determine the optimal scaling for initialization and learning rates of these unconventional layers.
Finally, we measure the scaling laws of different structures to compare how quickly their performance improves with compute. We propose a novel matrix family containing Monarch matrices, the Block Tensor-Train (BTT), which we show performs better than dense matrices for the same compute on multiple tasks. On CIFAR-10/100 with augmentation, BTT achieves exponentially lower training loss than dense when training MLPs and ViTs. BTT matches dense ViT-S/32 performance on ImageNet-1k with 3.8 times less compute and is more efficient than dense for training small GPT-2 language models.
\end{abstract}

\begin{figure*}[!t]
\centering
    \hspace{55mm}
    \includegraphics[width=0.8\linewidth]{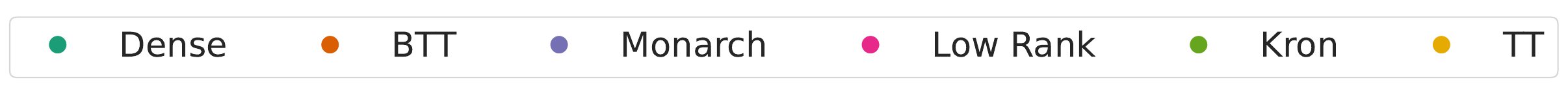} 
    \\
    \subfloat[Same compute, wider layers]{
    \includegraphics[height=0.23\linewidth]{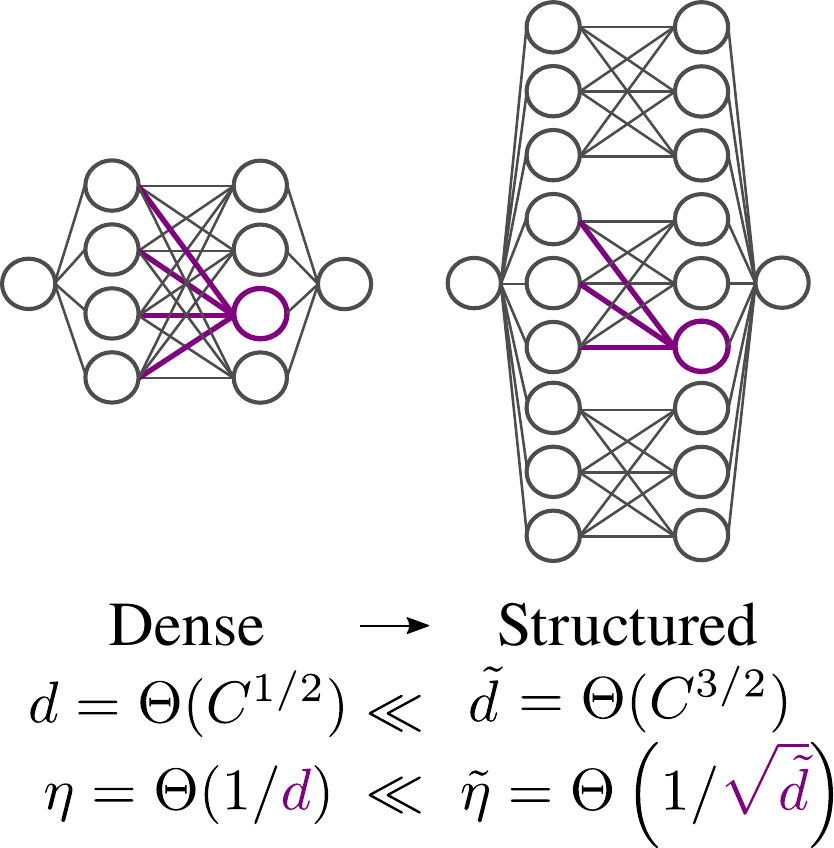}
    \label{fig:wider}
    \hspace{4mm}
    }
    \subfloat[CIFAR-10 train error]{
    \includegraphics[height=0.23\linewidth]{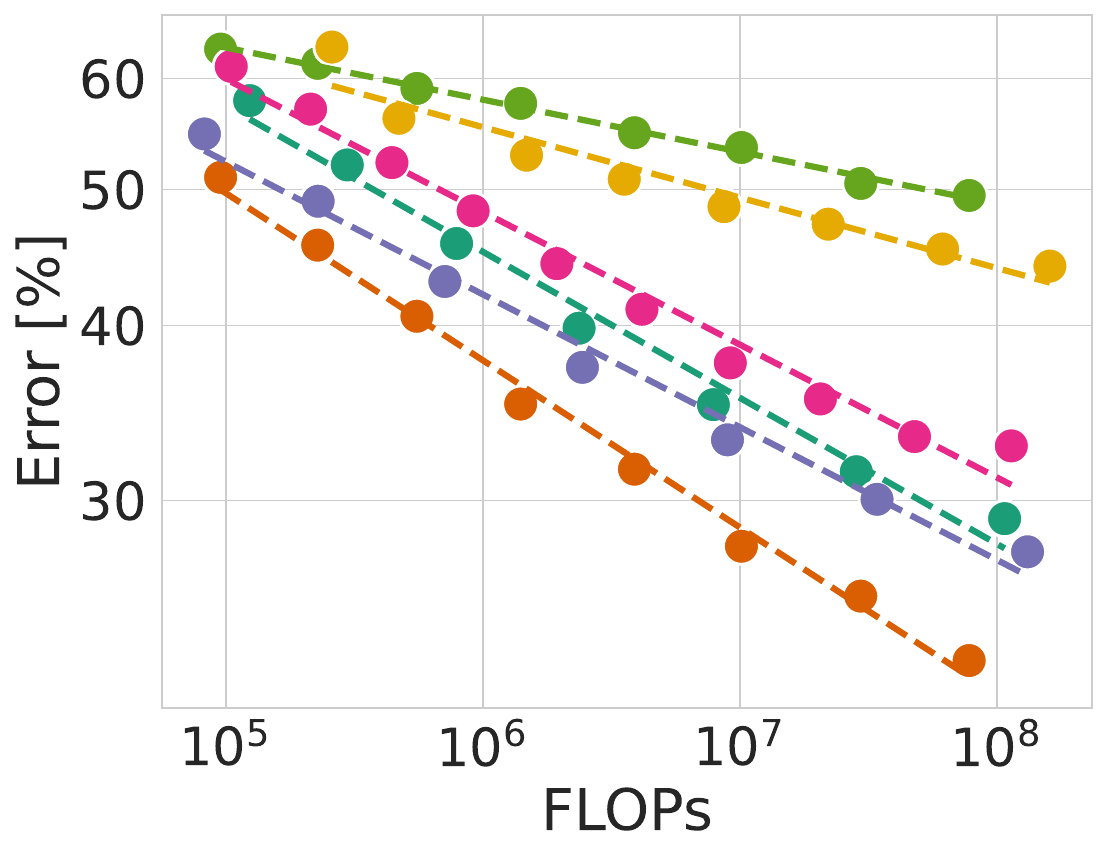}
    \label{fig:c10-train}
    }
    \subfloat[Structure-aware LR is crucial]{
    \includegraphics[height=0.23\linewidth]{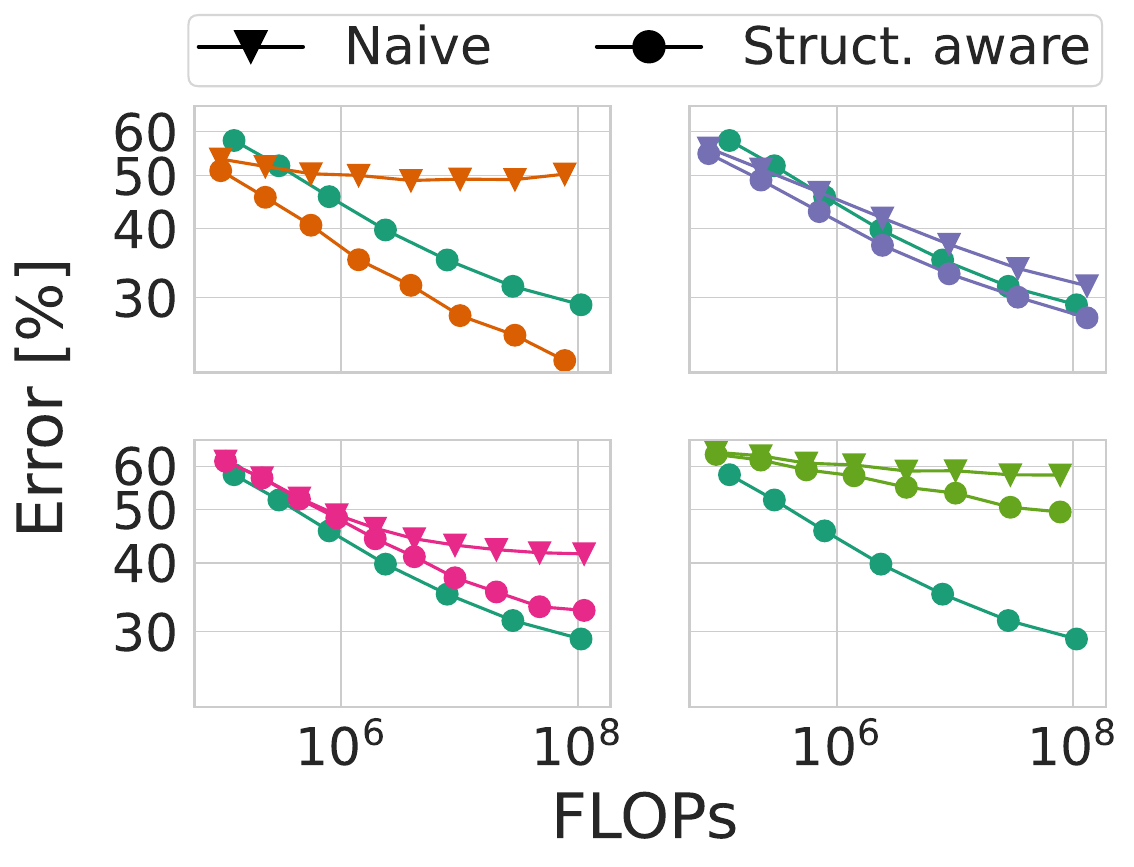}
    \label{fig:struct-aware-lr}
    }
   \caption{
     \textbf{Controlling for compute, replacing dense layers with structured matrices enables wider models and can lead to better performance.} (a) A neural network with structured matrices can be made much wider, but its learning rate needs to be scaled differently as a function of width since not all connections are present (\Cref{sec:sucessful}). The width $d$ of a dense layer scales as $C^{1/2}$ where $C$ is the compute per forward pass, while the width $\tilde{d}$ of a block diagonal layer is exponentially larger, scaling as $C^{2/3}.$ The optimal learning rate $\eta$ of the dense layer and $\tilde{\eta}$ of the block diagonal layer scales differently as $d^{-1}$ and $\tilde{d}^{-1/2}$ respectively. (b) Structured matrices can improve the training error scaling laws of MLPs on CIFAR-10 with data augmentation (\Cref{sec:experiments}). (c) Scaling the learning rate in a structure-aware fashion ($\bullet$) is crucial for performance (\Cref{sec:sucessful}), without which the benefit of structured layers does not emerge ($\blacktriangledown$).
     }
    \vspace{-4mm}
\end{figure*}

\section{Introduction} \label{sec:intro}
Regardless of their architectures, most neural networks consist of interleaved linear layers and simple non-linearities. In large foundation models such as GPT-3 \citep{brown2020language}, these linear layers consume the vast majority of the parameters and computation \citep{kaplan2020scaling}, and are primarily represented by dense matrices.
Substituting these dense matrices with structured matrices with fast matrix-vector multiplies (MVMs) has the potential to significantly improve the computational efficiency of these models. Unfortunately, there often isn't an obvious algebraic structure to exploit in the linear layers of such models, which process end-to-end learned token embeddings rather than objects with clear structures like images \citep{vaswani2017attention}.

Structured matrices, however, are not limited to encoding domain-specific inductive biases. They can also offer advantages over dense matrices by enabling different allocations of the same computational budget. For example, a structured layer can be much wider than a dense layer given the same number of parameters and compute. The compute cost $C$ of an MVM is $\order{d^2}$ for a $d \times d$ dense matrix, but only $\order{d^{3/2}}$ for a block diagonal matrix with $\sqrt{d}$ blocks. Consequently, given the same compute $C,$ the width can be at most $\order{C^{1/2}}$ for a dense layer, but $\order{C^{2/3}}$ for such a block diagonal layer. We can replace a dense layer of width $1024$ with a $10\times$ wider block diagonal layer, as illustrated in \Cref{fig:wider}.
Both layers have the same number of parameters and compute costs, but a larger width enables the model to potentially store more information in its activations and use more non-linearities to model complex functions. In this light, structured matrices do not merely approximate dense matrices but enable different ways of scaling up the models with compute that make them potentially more expressive.

To study how structured layers compare against dense layers as a function of compute, we will compare their scaling laws: how compute translates to performance as the models scale up. Across domains such as language, image, and video modeling, the loss or error rate $E$ of a well-trained neural network has shown to be highly predictable as a function of the compute $C$ required by the model, often well-described by a power law $E \propto C^{-\alpha}$ when data is not a bottleneck \citep{kaplan2020scaling, sharma2022scaling, hoffmann2022training}. If structured layers can achieve better scaling laws, they will outperform dense layers at scale, delivering exponentially better performance per unit compute if they can improve the scaling exponent $\alpha$.

In this work, we systematically study whether structured matrices can have better scaling laws than dense matrices, without relying on domain-specific algebraic structures so that our findings can apply to training foundation models broadly. 

\: \textbullet \, We show that structured layers often require drastically different learning rates and initialization scales compared to their dense counterparts, because their underlying trainable parameter matrices tend to be much smaller in size than the width of the layer (\Cref{fig:wider}). Naively using dense layer learning rates, structured layers often significantly underperform dense layers, as shown in \Cref{fig:struct-aware-lr}.

\: \textbullet \, Leveraging insights from \mup \citep{yang2023spectral} on how to optimally scale the initialization and learning rates for dense layers as a function of width, we show how to automatically determine the appropriate initialization and learning rate scales for structured linear layers. This structure-aware technique enables us to effectively train and scale a wide range of structured layers without additional tuning.

\: \textbullet \, We measure scaling laws for neural networks employing structured matrices as they scale, showing that structured layers can have better scaling exponents than dense matrices on some tasks. These results suggest that the scaling exponents are not necessarily determined solely by the task as previously hypothesized~\citep{bahri2021explaining,michaud2023quantization}.

\: \textbullet \, We identify matching parameter count to FLOPs\footnote{Here and elsewhere in the paper we use the more familiar term FLOPs as a stand-in for MACs (Multiply-Accumulate) operations to highlight when they match the number of parameters, even though $1$ MAC is technically $2$ FLOPs.} as a principle shared by the best-performing structures. Conversely, commonly used structures such as the Kronecker product and Tensor-Train decomposition violate this principle and underperform dense matrices in our experiments. Adhering to this principle can serve as important guidance for future work on designing more efficient linear layers.

\: \textbullet \, We introduce Block Tensor-Train (BTT) as a new family of expressive structured matrices, containing the Monarch matrices \citep{dao2022monarch} as a special case. The BTT family has better scaling laws than dense matrices on multiple tasks. On CIFAR-10/100 with augmentation, BTT achieves exponentially lower training loss than dense when training MLPs and ViTs. On ImageNet-1k, BTT matches dense ViT-S/32 performance with 3.8 times less compute. 

\: \textbullet \, We study divergences in training transformers with BTT layers, showing that weight normalization is required to avoid divergence due to unbounded growth of the activation.

We make our code available available \href{https://github.com/shikaiqiu/compute-better-spent}{\underline{here}}. We use the \texttt{Linear Operator} abstractions in
CoLA \citep{potapczynski2024cola} to prototype and compute efficient MVMs for structured matrices.

\begin{table*}[!t]
\centering
\begin{tabularx}{\textwidth}{l|c|c|l|X}
\toprule
\textbf{Structure} & \textbf{MVM FLOPs} & \textbf{\# Params} & \textbf{Modeling assumptions} & \textbf{Example applications} \\
\midrule
Dense & $d^{2}$ & $d^{2}$ & General linear maps & MLPs, Transformers \\
Low-Rank & $2rd$ & $2rd$ & Compression & Bottleneck layers, Linear attention \\
Convolution & $pd$ & $p$ & Translation equivariance & Images, Time-series \\
Kronecker & $2d^{3/2}$ & $2d$ & Sets, Graphs, Grids & GPs, Deep Sets, Attention, GNNs \\
Monarch & $2d^2/b$ & $2d^2/b$ & Flexible & Compute-efficient linear layers \\
TT & $2 r d^{3/2}$ & $2 r d$ & Subsystems, Local interactions & Hidden Markov Models, Spin systems \\
BTT & $2 r d^{3/2}$ & $2 r d^{3/2}$ & Flexible & Compute-efficient linear layers \\
\bottomrule
\end{tabularx}
\caption{\textbf{Overview of the computational properties, modeling assumptions, and applications of structured matrices we consider.} Some structures require the same FLOPs as parameters for a matrix multiply, while others require more FLOPs. $d$ is the size of the matrix, $r$ is the rank in low-rank, TT, and BTT, $p$ is the kernel size in a convolution, and $b$ is the number of blocks in Monarch. We assume 2 cores each of size $\sqrt{d}$ for Kronecekr, TT and BTT.}
\label{tab:structures_memory_compute}
\end{table*}

\section{Structured Alternatives to Dense Layers} 
\label{sec:preliminaries}

We now introduce the types of structured matrices we consider in this work. We review their computational properties and modeling assumptions, summarized in \Cref{tab:structures_memory_compute}. Without loss of generality, we consider $d \times d$ square matrices for notational simplicity.

\noindent \textbf{Low-rank.} \quad
A low-rank matrix can be parameterized as $\bm{W} = \bm{U} \bm{V}$
where $\bm{U} \in \mathbb{R}^{d \times r}$, $\bm{V} \in \mathbb{R}^{r \times d}$ and $r \leq d$ is its rank.
It has $2rd$ parameters and its MVM costs $2rd$ FLOPs. By first performing a dimension reduction on the input via $\bm{V}$, a low-rank matrix assumes that only a subspace of the input space is relevant to the task and is natural for compression \citep{zhao2024galore,wang2020linformer}. 

\noindent \textbf{Convolution.} \quad
Convolutions, or Toeplitz matrices, naturally model systems with translational symmetries such as images \citep{lecun98grad, krizhevsky2021alexnet, he2015resnet} and time-series \citep{wilson2013gaussian}. A convolution with kernel size $p$ has $p$ parameters and requires $\order{p d}$ FLOPs. Each parameter is used $\order{d}$ times in a convolution to impose translational symmetry. Alternatively, the Fast Fourier transform allows the convolution to be computed in $\order{d \log d}$ FLOPs.

\noindent \textbf{Kronecker.} \quad Kronecker product structure naturally arises in applications with structured data \citep{perez2017film, titsias2009variational, maron2020learning, saatcci2012scalable, wilson2015kernel}. A Kroncker product $\bm{W} =\bm{L} \otimes \bm{R}$ with $\bm{L} \in \mathbb{R}^{d_{1} \times d_{1}}$, $\bm{R} \in \mathbb{R}^{d_{2} \times d_{2}}$, $d = d_{1} \cdot d_{2},$ specifies a matrix whose MVM $\bm{y} = \bm{W} \bm{x}$ can be efficiently computed as $y_{\alpha\beta} = \sum_{\gamma} L_{\alpha\gamma} \sum_{\delta} R_{\beta\delta} x_{\gamma\delta},$ after reshaping the input $\bm{x}$ in row-major order into a $d_1 \times d_2$ matrix and followed by flattening $\bm{y}$ back to a vector.
Assuming $d_1 = d_2 = \sqrt{d}$, $\bm{W}$ has $2d$ parameters and requires $2 d^{3/2}$ FLOPs for an MVM. The Kronecker product uses each parameter $\sqrt{d}$ times, which can be made explicit by interpreting $\sum_{\delta} R_{\beta\delta} x_{\gamma\delta}$ (the same argument applies to the sum involving $\bm{L})$ as multiplying the vector $\bm{x}$ by a block-diagonal matrix $\bigoplus_{\gamma=1}^{\sqrt{d}} \bm{R}_\gamma,$ where all the blocks $\bm{R}_\gamma \in \R^{\sqrt{d} \times \sqrt{d}}$ are shared: $\bm{R}_\gamma = \bm{R}, \gamma = 1, \ldots, \sqrt{d}.$ This parameter-sharing naturally corresponds to the assumption that the input $\bm{x}$ represents a set of objects of the same kind, such as nodes in a graph \citep{kipf2016semi}, patches of an image \citep{tolstikhin2021mixer}, points on a grid \citep{saatcci2012scalable}, or words in a sentence \citep{vaswani2017attention,elhage2021mathematical}.

\noindent \textbf{Monarch.} \quad
Introduced in \citet{dao2022monarch}, a Monarch matrix
is defined as the product $\bm{P} \bm{L} \bm{P}^{\top} \bm{R}$ where $\bm{P}$ is a row-major to column-major permutation
and $\bm{L}, \bm{R}$ are two block-diagonal matrices: $\bigoplus_{\beta=1}^{\sqrt{d}} \bm{L}_\beta, \bigoplus_{\gamma=1}^{\sqrt{d}} \bm{R}_\gamma$. Monarch requires $2d^{3/2}$ FLOPs for an MVM and has $2d^{3/2}$ parameters. The efficient multiply for Monarch can be written as $y_{\alpha\beta} = \sum_{\gamma} L_{\alpha\tc{\beta}\gamma} \sum_{\delta} R_{\beta\tc{\gamma}\delta} x_{\gamma\delta},$ where $R_{\beta\tc{\gamma}\delta} = (\bm{R}_{\tc{\gamma}})_{\beta\delta}$ and $L_{\alpha\tc{\beta}\gamma} = (\bm{L}_{\tc{\beta}})_{\alpha\gamma}$ and we have colored the block dimensions $\tc{\beta}, \tc{\gamma}$. Monarch can be viewed as a relaxation of the Kronecker product where parameters that were shared across the block dimensions are now made independent. Monarch matrices do not make strong assumptions about the structure of the input. In practice, the number of blocks $b$ in $\bm{L}$ and $\bm{R}$ are often chosen to be much less than $\sqrt{d}$ to reduce sparsity \citep{dao2022monarch, fu2023mixer}. In this case, Monarch has $2d^2/b$ parameters and requires $2d^2/b$ FLOPs for an MVM.

\noindent \textbf{Tensor-Train.} \quad
The Tensor-Train (TT) decomposition \citep{oseledets2011tt} specifies a set of $c$ cores $\bm{G}^{(i)} \in \mathbb{R}^{r_{i} \times  m_{i} \times  n_{i} \times  r_{i-1}}$
for $i=1, \ldots, c$ where $d = \prod_{i}^{} m_{i} = \prod_{i}^{} n_{i},$
$r_{i} \in \mathbb{N}$ and $r_{0} = r_{c} = 1$.
For ease of notation, we will focus on $c=2$ with $m_1=m_2=n_1=n_2=\sqrt{d}, r_{1} = r$, $\bm{G}^{(1)} = \bm{R} \in \R^{r \times \sqrt{d} \times \sqrt{d}}, \bm{G}^{(2)} = \bm{L} \in \R^{\sqrt{d} \times \sqrt{d} \times r}$,  though we present the general case in Appendix \ref{app:general}.
With the input and output as reshaped as $\sqrt{d} \times \sqrt{d}$ matrices, a TT matrix is equivalent to a sum over $r$ Kronecker products indexed by $\sigma=1,\ldots, r$:
\begin{equation} \label{eq:full}
    \begin{split}
      y_{\alpha\beta} = \sum_{\gamma\sigma} L_{\alpha\gamma\sigma} \sum_{\delta} R_{\sigma\beta\delta} x_{\gamma\delta}.
    \end{split}
\end{equation}

By increasing $r,$ referred to as the TT-rank, TT becomes more expressive relative to the Kronecker product. When $r=d,$ it can represent any $d \times d$ dense matrix. TT has $2rd$ parameters and costs $2rd^{3/2}$ FLOPs for an MVM. Like Kronecker, TT shares parameters along the block dimensions $\beta,\gamma$ and therefore uses each parameter $\sqrt{d}$ times in an MVM. The TT structure is natural for modeling systems that decompose into subsystems with local pairwise interactions, such as quantum spin chains and hidden Markov models \citep{fannes1992finitely, critch2014algebraic}.

\noindent \textbf{Block Tensor-Train.} \quad
We propose a novel family of structured matrices called Block Tensor-Train (BTT) matrices, by removing the parameter-sharing along the block dimensions $\tc{\beta},\tc{\gamma}$ in the TT structure. In the two core ($c=2$) case, a BTT matrix of BTT-rank $r$ is defined by two parameter tensors $\bm{R} \in \R^{r \times \sqrt{d} \times \sqrt{d} \times \sqrt{d}}$ and $\bm{L} \in \R^{\sqrt{d} \times \sqrt{d} \times \sqrt{d} \times r}.$ Its MVM is given by
\begin{equation} \label{eq:btt}
    \begin{split}
      y_{\alpha\beta} = \sum_{\gamma\sigma} L_{\alpha\tc{\beta}\gamma\sigma} \sum_{\delta} R_{\sigma\beta\tc{\gamma}\delta} x_{\gamma\delta}.
    \end{split}
\end{equation}

In \Cref{app:general}, we study the expressiveness of BTT, present a simple algorithm for projection onto the BTT family, and show BTT with rank $r=\sqrt{d}$ can represent any dense matrix (in constrast to $r=d$ for TT) when $c=2$ and analogous results for $c > 2$. Therefore, by varying the BTT rank, we effectively interpolate between Monarch matrices and dense matrices.

We use the \texttt{Linear Operator} abstractions available in \texttt{CoLA} \citep{potapczynski2024cola} to compute MVMs for these structures efficiently. In \Cref{app:runtime}, we show the structures we consider have asymptotically the same MVM runtimes as dense matrices as a function of FLOPs because they can be implemented through the same dense matrix multiply primitives, though they introduce non-trivial overhead for small matrix sizes with our current implementation.

\section{Optimizing Structured Matrices} \label{sec:sucessful}

To study the performance and scaling laws of unconventional layers, we must determine how to optimize them effectively by choosing appropriate initialization and learning rates as the models scale. As \Cref{fig:struct-aware-lr} illustrates, the optimal settings for structured matrices can differ significantly from dense matrices. We develop a technique based on the Maximal Update Parameterization (\mup) \citep{yang2021infty, yang2023iv, yang2021v} to automatically determine the optimal initialization and learning rate scaling for a generic structured layer given its structure and size, enabling us to train and scale various structured layers with good hyperparameters and minimal tuning. We focus on the Adam optimizer \citep{kingma2015adam} but discuss extensions to other optimizers in \Cref{app:other_opt}.

\subsection{Maximal Update Parameterization}
The Maximal Update Parameterization (\mup) \citep{yang2021infty, yang2023iv, yang2021v} specifies how to scale the initialization and learning rate of neural networks as their widths increase while maximizing feature learning in every layer \citep{yang2021infty}. \citet{yang2023spectral} provides an elementary derivation based on the spectral norm, which we now review.

In \mup, initialization and learning rates are chosen so that entries of each layer's output have size $\Theta(1)$ and are updated at a rate of $\Theta(1)$ per step throughout training. Here, big-$\Theta$ notation denotes scaling in the layer's width, omitting dependence on other quantities. If these conditions do not hold, the layer's output or update will either diverge or vanish for sufficiently large widths.
For a dense matrix $\bm{W} \in \R^{\dout \times \din},$ input $\bm{x} \in \R^\din,$ output $\bm{h} = \bm{W} \bm{x} \in \R^\dout,$ and output update $\Delta \bm{h} = \Delta\bm{Wx}$ due to a weight update $\Delta\bm{W}$, \mup requires $\norm{\bm{h}}_2 = \Theta(\sqrt{\dout})$ and $\norm{\Delta\bm{h}}_2 = \Theta(\sqrt{\dout}).$ During training, gradient descent aligns $\bm{x}$ with the top singular subspace of $\bm{W}$ and $\Delta\bm{W}$ \citep{yang2023spectral,yang2023iv}, so $\norm{\bm{h}}_2 = \Theta(\norm{\bm{W}}_2 \norm{\bm{x}}_2)$ and $\norm{\Delta\bm{h}}_2 = \Theta(\norm{\Delta\bm{W}}_2 \norm{\bm{x}}_2).$ Assuming $\bm{x}$ is entry-wise  $\Theta(1)$, we want $\norm{\bm{W}}_2 = \Theta(\sqrt{\dout/\din})$ and $\norm{\Delta\bm{W}}_2 = \Theta(\sqrt{\dout/\din}).$ To ensure the desired spectral norm at initialization, entries of $\bm{W}$ are drawn from $\mathcal{N}(0, \sigma^2)$ with $\sigma = \Theta(\sqrt{\min(\din, \dout) / d^2_\mathrm{in}})$. For the updates, the gradient $\nabla_{\bm{W}} \mathcal{L} = \frac{1}{B}\sum_{i=1}^{B} \nabla_{\bm{h}_i}\mathcal{L} \cdot \bm{x}_i^\top$ has $\Theta(1)$ stable rank, assuming the batch size $B$ is constant, so its spectral norm scales the same way as its Frobenius norm. Since Adam normalizes the gradient to be entry-wise $\Theta(1),$ the normalized gradient has Frobenius norm $\Theta(\sqrt{\din \dout})$. Therefore, an Adam learning rate of $\Theta(1 / \din)$ ensures the desired spectral norm.

Once the optimal learning rate $\eta^*$ is found for a particular width $\din,$ it can be transferred to any other width $d'_\mathrm{in}$ by setting the new learning rate as $\eta^* \cdot \frac{\din}{d'_\mathrm{in}},$ assuming $\din$ and $d'_\mathrm{in}$ are sufficiently large \citep{yang2021v}. For architectures, \mup deviates from conventional initializations mainly in the last layer, where $\sigma = \Theta(1/\din)$ according to \mup but $\sigma = \Theta(\sqrt{1/\din})$ according to more conventional strategies \citep{lecun2002efficient, glorot2010understanding, he2015delving}.

\begin{figure*}[!t]
\centering
    \includegraphics[width=0.38\linewidth]{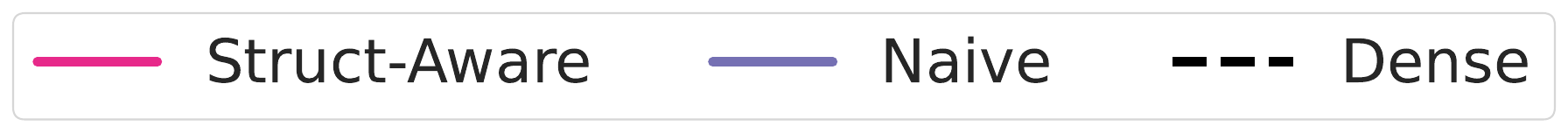}
    \hspace{10mm}
    \includegraphics[width=0.45\linewidth]{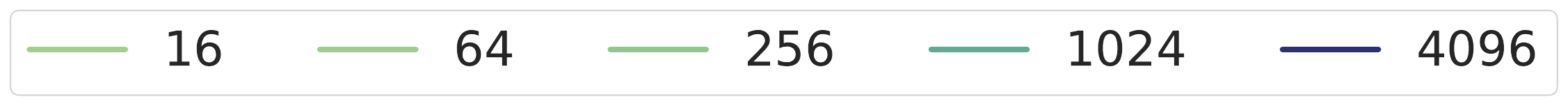}
    \\
    \subfloat[Stable feature learning]{
    \includegraphics[height=0.23\linewidth]{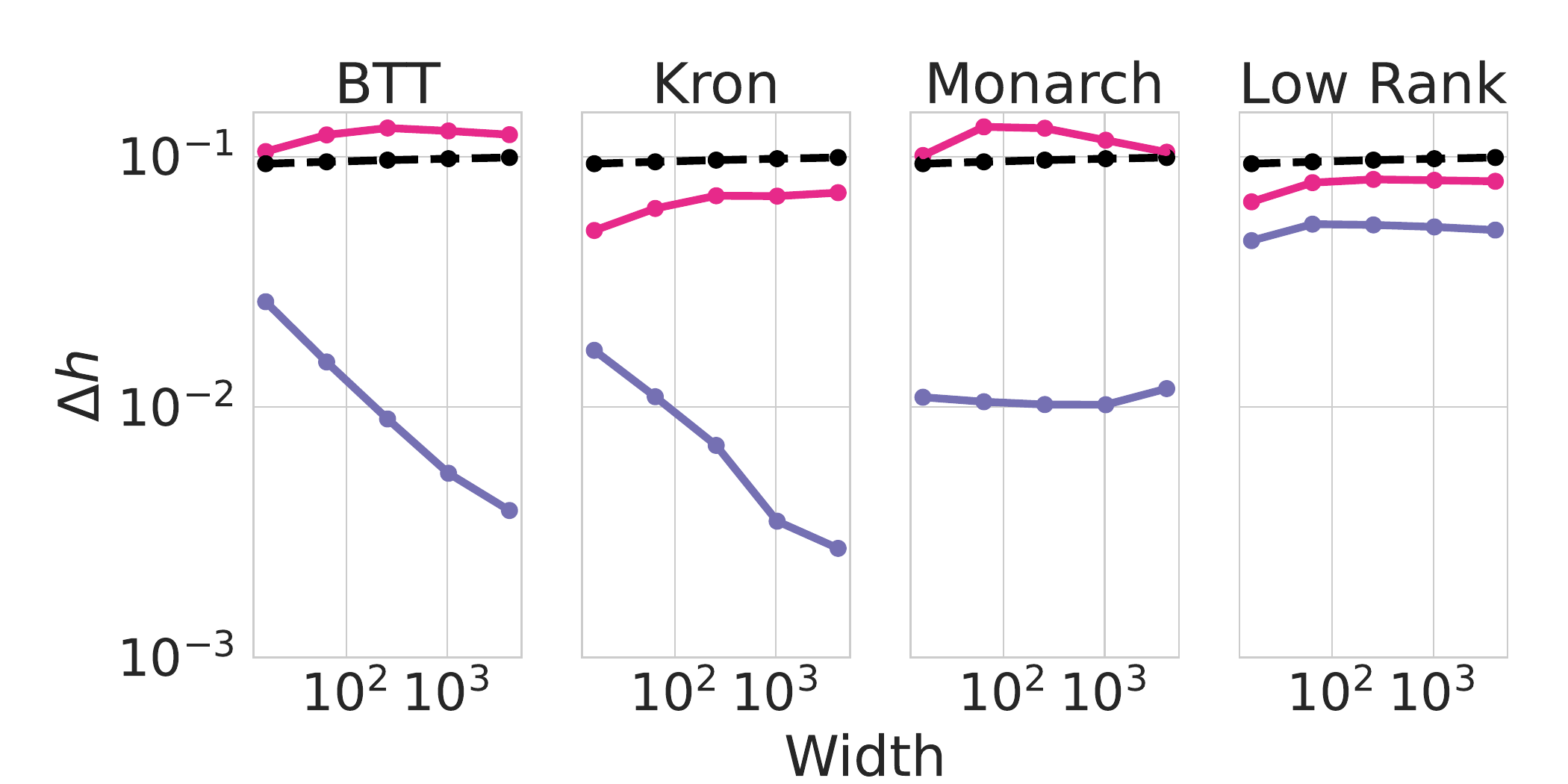}
    \label{fig:feature_update}
    }
    \subfloat[Stable optimal learning rate]{
    \includegraphics[height=0.23\linewidth]{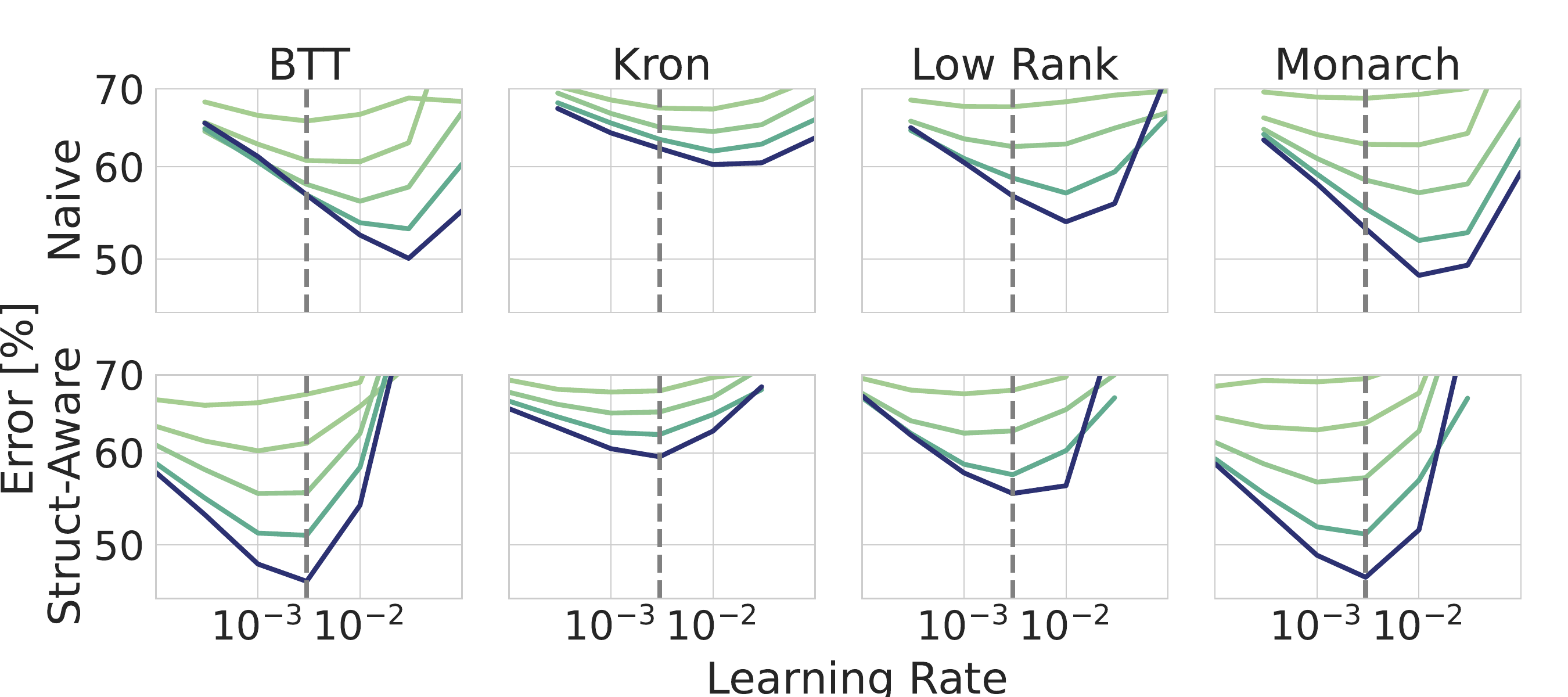}
    \label{fig:lr_land}
    }
   \caption{
   \textbf{Structure-aware learning rate scaling results in stable feature learning and stable optimal learning rate as we vary the structure and model size}. (a) The RMS of the changes $\Delta h$ of the last layer features is stable as the models are scaled up in width, but is smaller or vanishes if we naively use the learning rate for the dense model. (b) The optimal learning rate is stable as we vary the structure and width, provided we use structure-aware learning rates. Here we use Monarch with 16 blocks.
   }
    \label{fig:act}
    \vspace{-5mm}
\end{figure*}

\begin{figure}[!b]
\centering
    \includegraphics[width=0.8\linewidth]{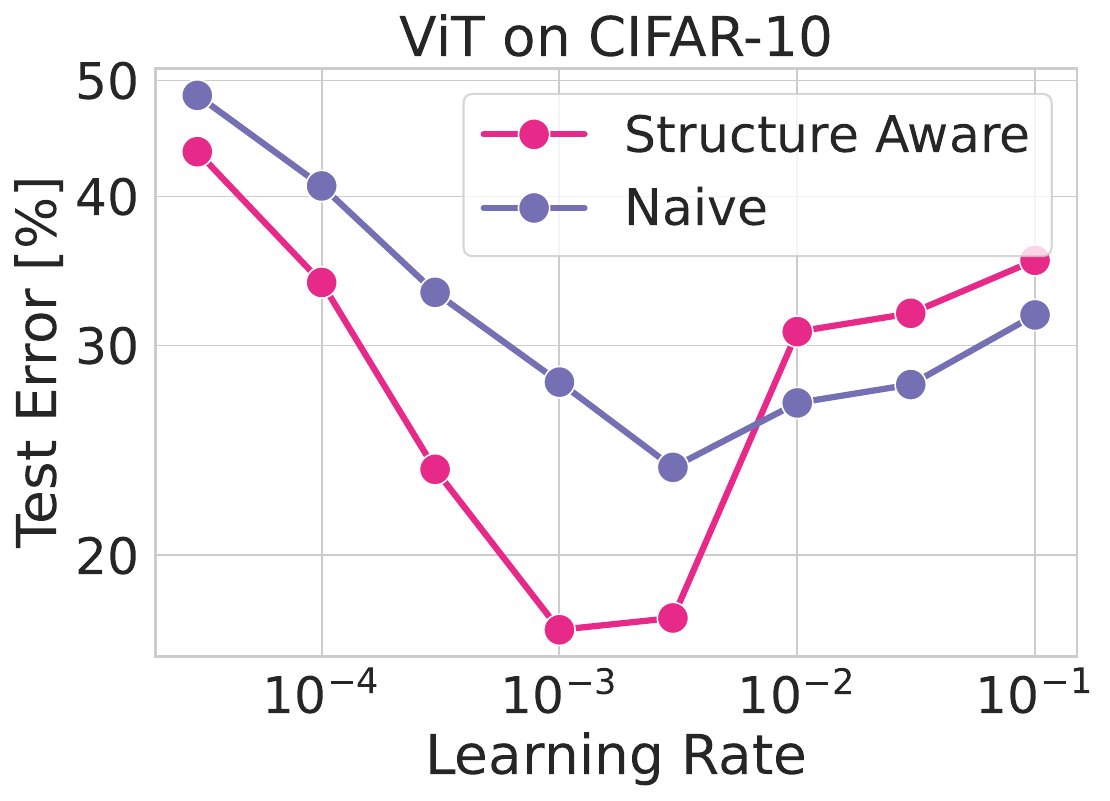}\\
   \caption{
   \textbf{Structure-aware learning rates improve performance even after tuning the learning with grid search.}
   Test error of ViT ($d=1024$) on CIFAR-10 where the feed-forward layers are replaced using BTT. 
   }
    \label{fig:ffn}
    \vspace{-4mm}
\end{figure}

\subsection{Identifying \mup for Structured Matrices} \label{sec:struct-aware}
The above scaling of learning rate and initialization assume dense matrices and don't immediately carry over to arbitrarily structured matrices. For example, for a Kronecker product $\bm{W} = \bm{L} \otimes \bm{R}$ where $\bm{W} \in \R^{d\times d}$ and $\bm{L}, \bm{R} \in \R^{\sqrt{d} \times \sqrt{d}},$ one intuitively expects that the optimal learning rates for parameters $\bm{L}$ and $\bm{R}$ in this layer to scale as $\Theta(1/\sqrt{d}),$ the size of the actual learnable parameter matrices, rather than naively as $\Theta(1/d)$ based only on the width of the layer.

Since many structured matrices are ultimately compositions of smaller dense matrices and fixed, norm-preserving linear transformations (e.g. reshapes), as exemplified in \Cref{sec:preliminaries}, we can decompose the problem by applying the same spectral considerations to each dense component separately, effectively treating each structured layer as a deep linear network. Suppose the MVM $\bm{W}\bm{x}$ can be computed as $\bm{W} \bm{x} = \bm{G}_k \bm{P}_{k} \ldots \bm{G}_1 \bm{P}_1 \bm{x}$ where each $\bm{P}_i$ is a fixed, norm-preserving linear transformation, such as the product of a permutation and a reshape,
and multiplication by $\bm{G}_i$ denotes a batched MVM, i.e.,
$(\bm{G}_i \bm{x})_{b \mu} = \sum_{\nu} (G_i)_{b\mu\nu} x_{b\nu}$ for some dense tensor $\bm{G}_{i} \in \R^{B_i \times d_{\mathrm{out}}^{i} \times d_{\mathrm{in}}^{i}},$ where $b$ is an abstract batch-like dimension.
Then to ensure that the activations have size $\Theta(1)$ and all parameters are updated as much as possible to maximize feature learning \citep{yang2023spectral}, we require the initialization and updates to each slice $(\bm{G}_i)_b \in \R^{d_{\mathrm{out}}^{i} \times d_{\mathrm{in}}^{i}}$ of
$\bm{G}_i$ to have $\Theta\Bigl(\sqrt{d_{\mathrm{out}}^{i} / d_{\mathrm{in}}^{i}}\Bigr)$ spectral norm.
Thus we initialize each $\bm{G}_i$ with standard deviation
$\Theta\Bigl( \sqrt{\min(d_{\mathrm{in}}^{i}, d_{\mathrm{out}}^{i}) / (d_{\mathrm{in}}^{i})^{2}} \Bigr)$
and set its Adam learning rate as $\Theta\left(1 / d_{\mathrm{in}}^{i}\right)$. When used in the last linear layer in a residual block, we zero-initialize the last component $\bm{G}_{k}$, which is compatible with \mup by setting the hidden constant in $\Theta(\cdot)$ to 0 \citep{yang2021v}.

\noindent \textbf{Transferring learning rate between structures.} \quad Once the optimal learning rate $\eta^*$ is known for a $\dout \times \din$ dense layer, we can infer the optimal learning rate $\eta_i^*$ of each component $\bm{G}_i$ of the corresponding structured layer as $\eta_i^* = \kappa_i \cdot \eta^*,$ where $\kappa_i = \frac{\din}{d_{\mathrm{in}}^{i}} \cdot \delta_i$ for some constant $\delta_i.$ Here $\frac{\din}{d_{\mathrm{in}}^{i}}$ accounts for the $\Theta(\mathrm{width}^{-1})$ scaling of optimal learning rate prescribed by \mup, with width identified with $\din$ and $d_{\mathrm{in}}^{i}$ respectively for the dense matrix and $\bm{G}_i$, and $\delta_i$ accounts for potential differences in the constants omitted by $\Theta(\cdot)$ for the dense matrix and $\bm{G}_i$. While the precise value of $\delta_i$ is not theoretically determined by \mup, we adopt the heuristic $\delta_i = 1/k$ where $k$ is the number of learnable dense components so that the overall updates to the output of this layer is roughly preserved, since $\Delta \bm{h}$ has $k$ leading order terms:
\begin{equation}
\begin{aligned}
    \Delta \bm{h} &= \sum_{i=1}^{k} \bm{G}_k \bm{P}_{k} \ldots
\Delta \bm{G}_i \bm{P}_{i} \ldots \bm{G}_1 \bm{P}_1 \bm{x} \\
 & + \order{\Delta \bm{G}^2}.
\end{aligned}
\end{equation}

$\delta_i$ can be further tuned empirically around $1/k$ to maximize performance, though we will show the $1/k$ heuristic is sufficently good in practice.

As $\bm{G}_i$'s are often much smaller in size than the matrix $\bm{W}$ it parameterizes, the required learning rate multiplier $\kappa_i$ is often a large number.
For example, suppose we initially represent $\bm{W} \in \mathbb{R}^{d \times d}$ as a dense matrix and find $\eta$ is an effective learning rate during training. If we now instead represent $\bm{W} = \bm{L} \otimes \bm{R}$ where $\bm{L}, \bm{R} \in \R^{\sqrt{d} \times \sqrt{d}},$
we would then need to scale up the learning rate for both $\bm{L}, \bm{R}$ by a factor of $\Theta(\sqrt{d}),$ which grows arbitrarily large for large $d.$ We show the Adam learning rate multipliers required for various structures in \Cref{tab:structures_mup}, adopting our heuristic of $\delta_i = 1/k$.

\subsection{Empirical Validation}
We now empirically validate the effectiveness of our structure-aware learning rate scaling.
We compare it to the naive, structure-agnostic approach that parameterizes the learning rate $\eta_i$ for each parameter tensor $\bm{G}_i$ in a $\dout \times \din$ structured layer as $\eta_i = \eta_0 \frac{d_0}{\din} \propto 1/\din,$ where the base learning rate $\eta_0$ and the base width $d_0$ are constants, corresponding to scaling the learning rate optimally according to \mup if the layer were dense. 
The structure-aware approach additionally applies the structure-dependent learning rate multipliers $\kappa_i$ in \Cref{tab:structures_mup} so that $\eta_i = \eta_0 \frac{d_0}{\din} \kappa_i.$ We use $d_0 = 64$ throughout this section.

\noindent \textbf{Stable feature learning.} We train an MLP with 2 hidden layers without bias on CIFAR-10 with width $d \in \{16, 64, 256, 1024, 4096\}$ and a base learning rate $\eta_0 = 3 \cdot 10^{-3}$.
For a given width, we track the root mean square (RMS) of $\Delta \bm{h}_{t} = \bm{h}_{t+1} - \bm{h}_{t}$ at every step $t$,
where $\bm{h}_{t} \in \mathbb{R}^{d}$ is the activation of the last layer before the classification head.
We then plot the the average RMS over $500$ steps for different widths and structures.
As seen in Figure \ref{fig:act}, structure-aware learning rate scaling produces consistent feature learning for all structures used with no tuning. In contrast, the naive approach causes much smaller or vanishing updates to the features. The effect is most pronounced for BTT and Kronecker, for which $\kappa_i$ grows without bound for both $\bm{L}$ and $\bm{R}$ as the width increases.

\noindent \textbf{Stable optimal learning rate.} We test if the structure-aware learning scaling preserves the learning rate landscape for all structures so that once an optimal learning rate is found for the dense model with some width, it can be directly transferred to all other structures and widths. We train a 2-layer MLP on CIFAR-10 with augmentation (see \Cref{sec:experiments} for details) for 100 epochs, using a base learning rate of $3 \cdot 10^{-3},$ the optimal value for a dense model at with $d_0=64$. In the first row of Figure \ref{fig:lr_land}, we show the train error as a function of the base learning rate $\eta_0$ when scaled to other widths and structures using the naive approach, which is optimal for the dense model but clearly not for the other structures. By contrast, in the second row, the structure-aware approach approximately stabilizes the learning rate landscape across structures and widths, significantly reducing the cost for exploring different structures. Slight deviation at small widths is expected because the optimality of \mup relies on convergence to the infinite-width limit \citep{yang2021infty}.

\noindent \textbf{Improved performance even after tuning.} Finally, we show in \Cref{fig:struct-aware-lr} the performance of structured models quickly saturate as they are scaled up without structure-aware learning rates. Monarch is an exception, for which the multipliers in \Cref{tab:structures_mup} are closer to 1 because we use $b=4$. In this case, the learning rate multiplier required for Monarch is only $2$ and independent of scale, which may explain why \citet{dao2022monarch} still achieves good performance with Monarch by reusing the dense learning rates.

Furthermore, the structure-aware approach not only reduces the tuning cost for structured layers, but is necessary for optimal performance if the structures differ across layers, even when we perform a grid search over the base learning rate $\eta_0$. Consider a transformer of hidden dimension $d$ where only the feed-forward layers (FFN) are replaced with BTT and the attention projection matrices are dense. Since the optimal learning rate is $\Theta(1/\sqrt{d})$ for the FFN layer but $\Theta(1/d)$ for the attention projection, the naive approach would have to choose between using a learning rate too large for the attention projection or a learning rate too small for the FFN, whereas the structure-aware approach does not have this problem. In Figure \ref{fig:ffn}, we show that for a ViT with BTT-structured FFNs, the structure-aware approach indeed achieves much better performance even if we tune the base learning rate.

\begin{figure*}[!t]
\vspace{0mm}
\centering
    \includegraphics[width=0.7\linewidth]{figs/mlp_cifar10_legend.pdf}\\
    \subfloat[MLP train error and exponent]{
    \includegraphics[height=0.18\linewidth]{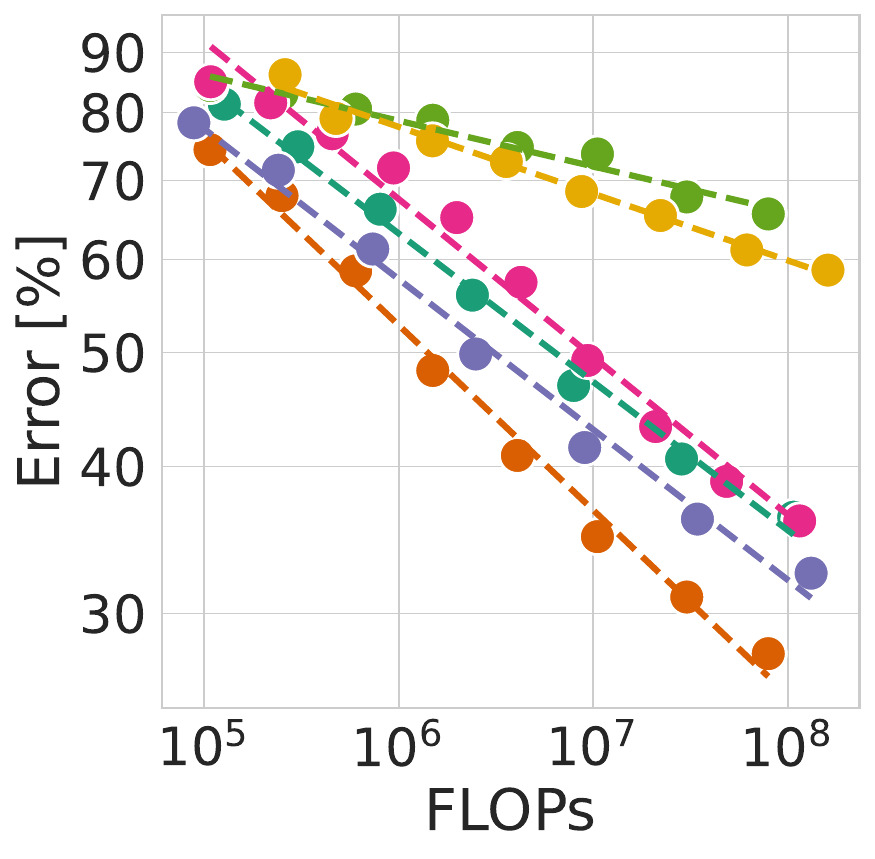}
    \includegraphics[height=0.18\linewidth]{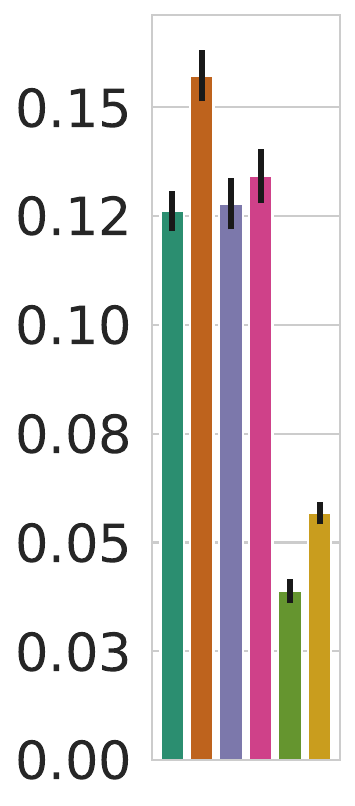}
    }
    \subfloat[MLP test error]{
    \includegraphics[height=0.18\linewidth]{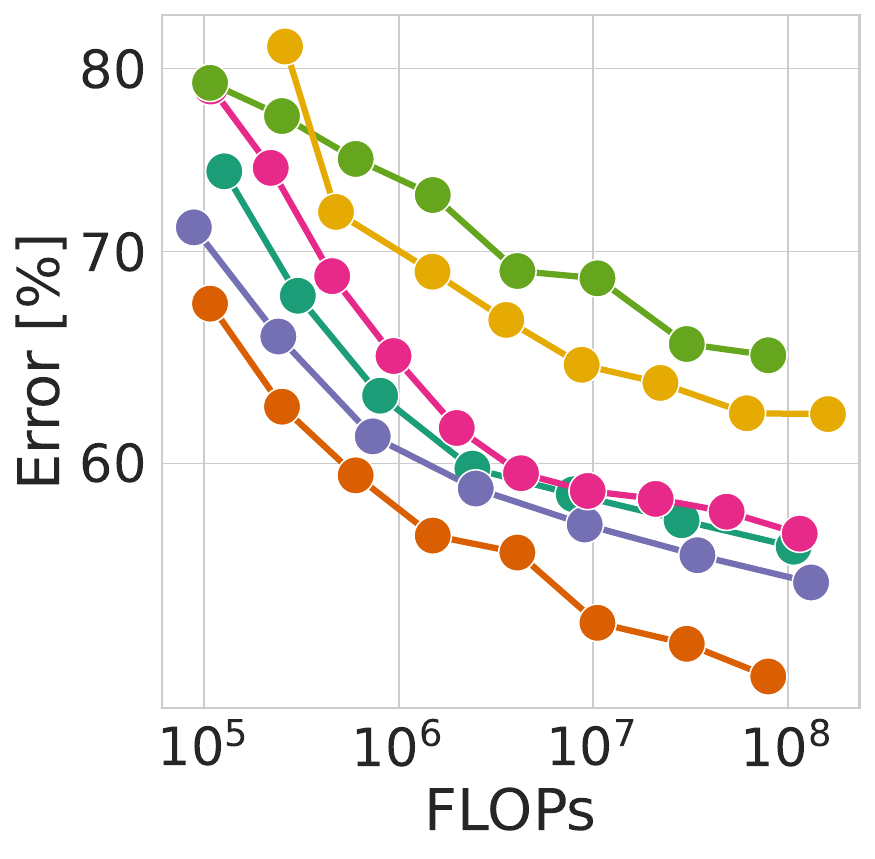}
    }
    \subfloat[ViT train error and exponent]{
    \includegraphics[height=0.18\linewidth]{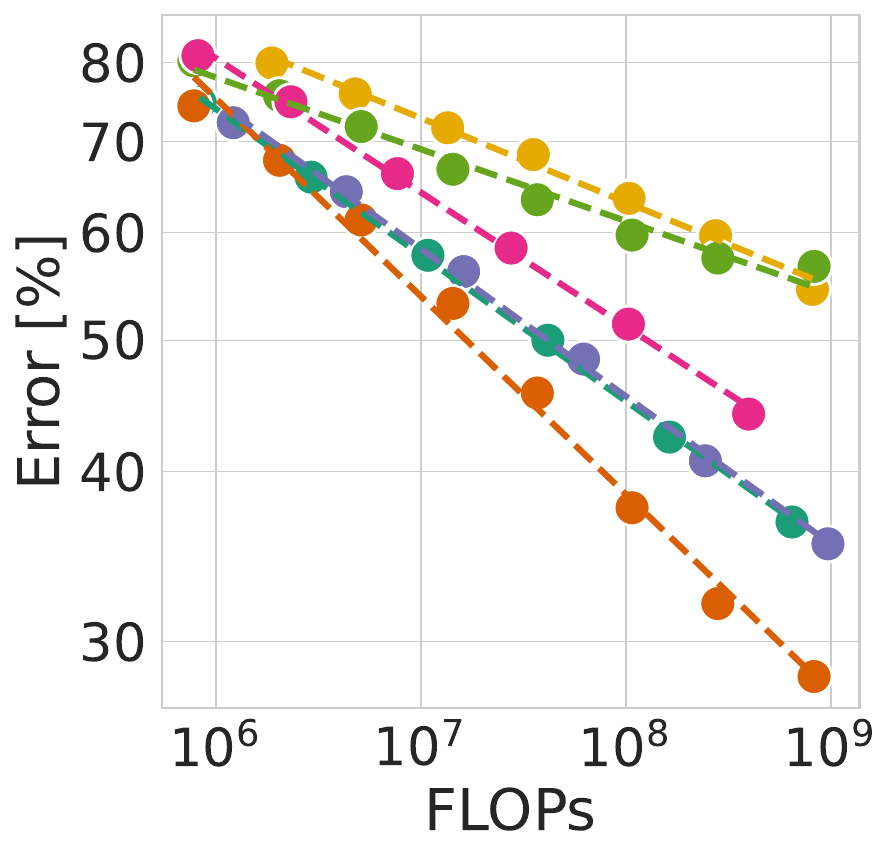}
    \includegraphics[height=0.18\linewidth]{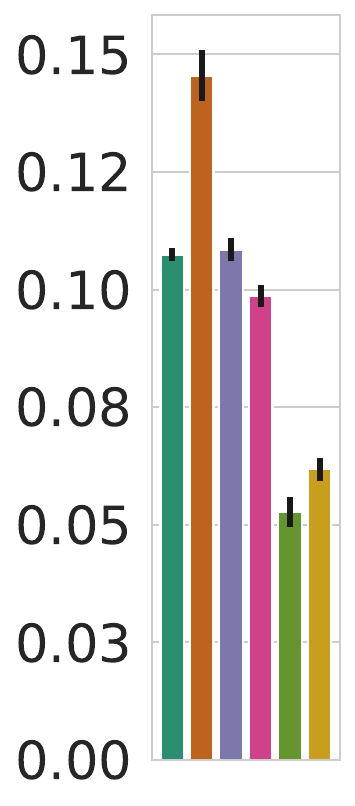}
    }
    \subfloat[ViT test error]{
    \includegraphics[height=0.18\linewidth]{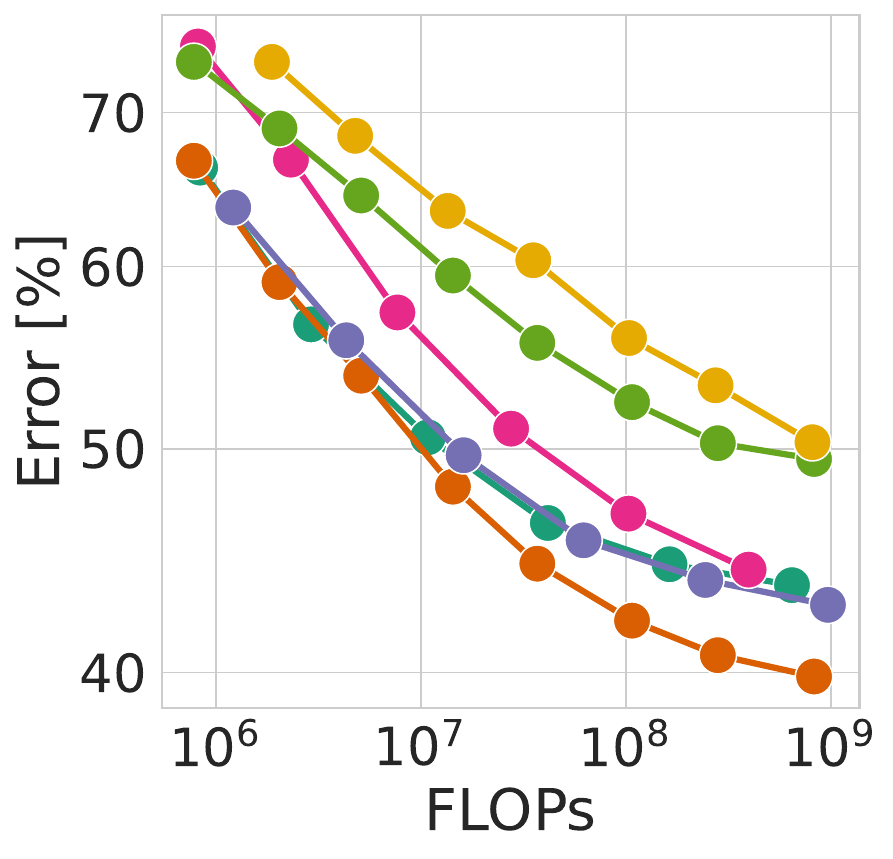}
    }
   \caption{
     \textbf{Using structured matrices changes the scaling laws of MLPs and ViTs trained on CIFAR-100.} We find 1) BTT achieves the best scaling, and 2) structures with FLOPs equal to parameters scale better than those with parameter sharing (Kronecker and TT)
     }
    \label{fig:struct-extended}
    \vspace{-4mm}
\end{figure*}

\section{Scaling Laws of Structured Matrices} \label{sec:experiments}
Having developed an effective procedure to automatically scale the initialization and learning rates for structured layers, we now aim to understand how various structures compare in performance.

When data is not a bottleneck, a neural network's test error or loss on a task follows a power law $E \propto P^{-\alpha_P}$ if trained to (near) convergence, where $P$ is the number of parameters and $\alpha_P$ is a constant \citep{kaplan2020scaling, hoffmann2022training, henighan2020scaling}. For dense models, compute per forward pass $C \propto P$, so $E \propto C^{-\alpha_C}$ for some constant $\alpha_C$. We explore how different structures change how $E$ scales with $C$, as $P$ does not consistently relate to training or inference cost when varying the structure (\Cref{tab:structures_memory_compute}).

We train all models for a fixed number of iterations $T$, so the total training compute $C_\mathrm{tot} \propto C$. Thus, the scaling laws in $C$ can differ from compute-optimal scaling laws, which require carefully optimizing the allocation of $C_\mathrm{tot} \propto C T$ between $C$ and $T$ \citep{kaplan2020scaling, hoffmann2022training}, which we leave to future work.

To compare multiple structures across compute scales, we conduct experiments primarily using MLPs and ViTs on CIFAR-10 and CIFAR-100. In \Cref{sec:trans}, we present larger-scale experiments on ImageNet and language modeling. With limited training data in CIFAR-10 and CIFAR-100, we apply heavy augmentation to alleviate over-fitting. The augmented training set is sufficiently large, resulting in relatively clean power-law scaling of training error with $C$. We extract these power law parameters, reflecting the expressivity afforded by each structure as a function of $C$, and visualize the scaling of test error with $C$, which is not well-described by a power law due to train-test discrepancy.

\noindent \textbf{Experimental setup.} \quad We use CIFAR-10 and CIFAR-100 datasets, applying random crop, random flip, MixUp ($\alpha_\mathrm{mixup}=0.8$) augmentations, and label smoothing of $0.3$, following \citet{bachmann2023scaling}. We use the same MLP architecture as in \citet{bachmann2023scaling}, but apply a fixed random permutation to the pixels before feeding them to the MLP so our results will more likely generalize to non-image data. We also use ViTs \citep{dosovitskiy2022vit} with $8\times8$ patches. We train MLPs for 500 epochs with batch size of 1024, and ViTs for 200 epochs with batch size of 256. To scale up the model, we increase its width while holding the depth constant.
For structured models, we replace all except the classification layer with structured layers, though we keep the input layer dense for low rank to avoid an information bottleneck at the first layer. For Monarch, we set the number of blocks $b=4$ following  \citet{dao2022monarch} unless stated otherwise. We use BTT with two cores and various BTT-ranks. Further experiment details are in \Cref{app:exp-details}.

\noindent \textbf{Scaling exponents are structure-dependent.} \quad In \Cref{fig:c10-train} and \Cref{fig:struct-extended}, we find the training error $E$ has an approximate power law relation to the compute $C:$ $E \propto C^{-\alpha_C}$, for both MLPs and ViTs, where the exponent $\alpha_C$ varies significantly among structures. We show the best-fit exponent $\alpha_C$ and its standard error for each structure and plot the fitted power law trends.
Monarch ($b=4$) achieves equal or lower train and test error than dense for the same amount of compute, though it does not improve the scaling exponent of training error. BTT has the largest scaling exponent and consistently outperforms all other structures. We use BTT with two cores and rank 1, equivalent to a Monarch with $\sqrt{d}$ blocks, but BTTs with higher ranks also improve scaling as we will soon show.

\noindent \textbf{Parameters equal FLOPs leads to better scaling laws.} \quad Figure~\ref{fig:c10-train} and Figure~\ref{fig:struct-extended} reveal a qualitative difference between the scaling behavior of structures that perform parameter-sharing, i.e. Kronecker and TT, and those that do not, having parameters equal to FLOPs. Structures that do not share parameters are more flexible per unit of compute, and consistently achieve better scaling laws.

Recent works proposing to explain scaling laws from the data manifold dimension \citep{bahri2021explaining, sharma2022scaling} can naturally explain worse scaling exponents due to parameter-sharing. This theory predicts the scaling exponent $\alpha_P$ with respect to parameters is determined only by the intrinsic dimension of the data manifold, explaining why architectural details often only have minor impacts on the scaling laws \citep{kaplan2020scaling}. If changing the matrix structure leaves $\alpha_P$ invariant, then the scaling exponent $\alpha_C$ will depend on the structure in a simple way: if $C \propto P^\beta,$ then $\alpha_C = \alpha_P/\beta,$ that is, the more parameters sharing, the smaller the exponent $\alpha_C$. For example, $\beta = 1$ for dense, low-rank, and BTT, but $\beta = 3/2$ for Kronecker and TT. However, this exact factor underestimates the observed differences in the exponents between Kronecker, TT, and dense, and does not explain why BTT has a larger exponent. 
A more accurate model is needed to explain the observed structure-dependence of the scaling exponents.

\begin{figure}[!t]
\centering
    \subfloat[BTT]{
    \includegraphics[width=0.5\linewidth]{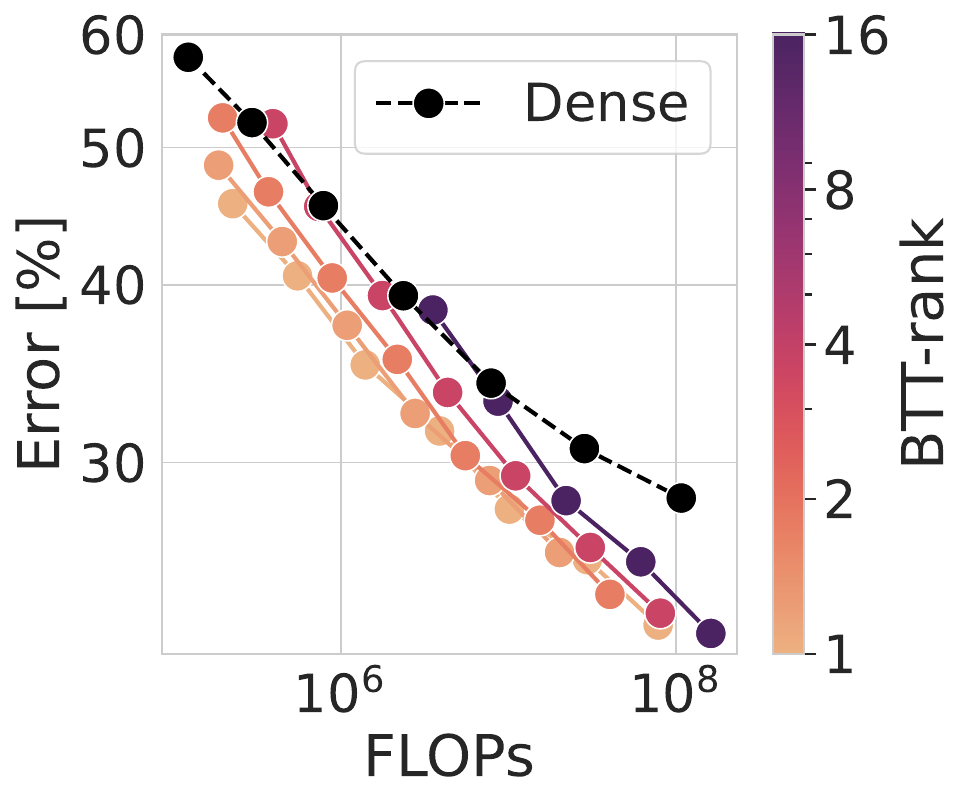}
    \label{fig:btt_rank_flops}
    }
    \subfloat[Monarch]{
    \includegraphics[width=0.5\linewidth]{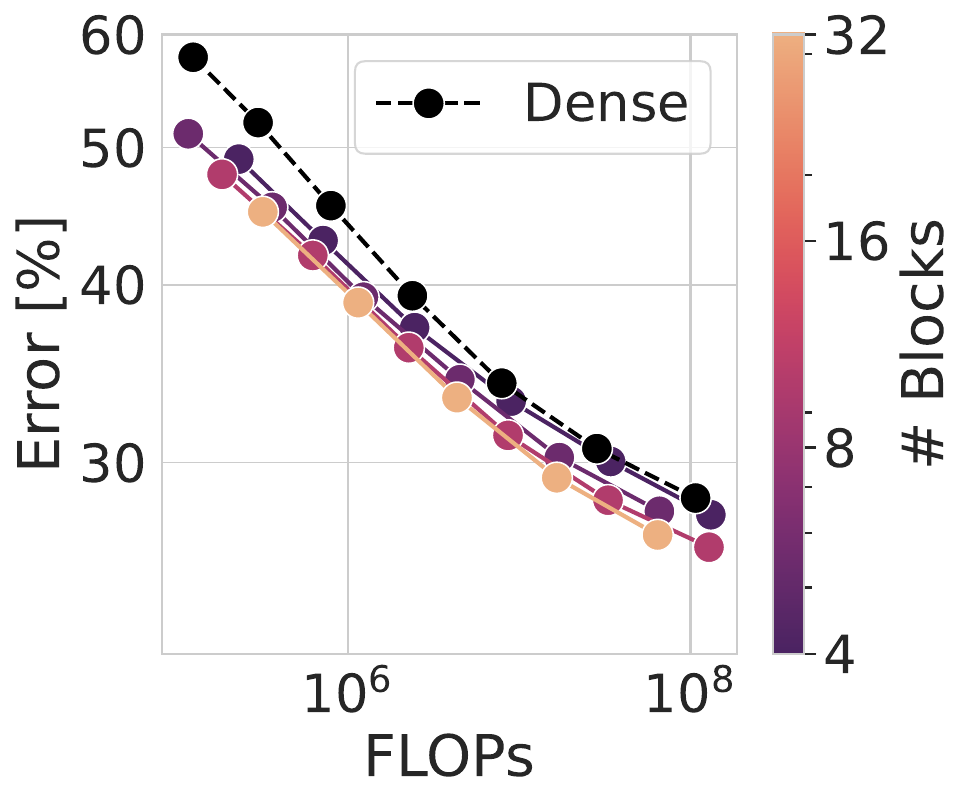}
    \label{fig:blocks_flops}
    }
   \caption{ \textbf{Less compute per dimension is more compute-efficient on CIFAR-10.}
   (a) BTT with a lower rank achieves lower train error per FLOP.
   (b) Monarch with more blocks achieves lower train error per FLOP. A lighter color indicates less compute per dimension. 
   }
    \label{fig:compute_per_dim}
    \vspace{-4mm}
\end{figure}

\begin{figure}[!t]
\centering
    \subfloat[BTT]{
    \includegraphics[width=0.5\linewidth]{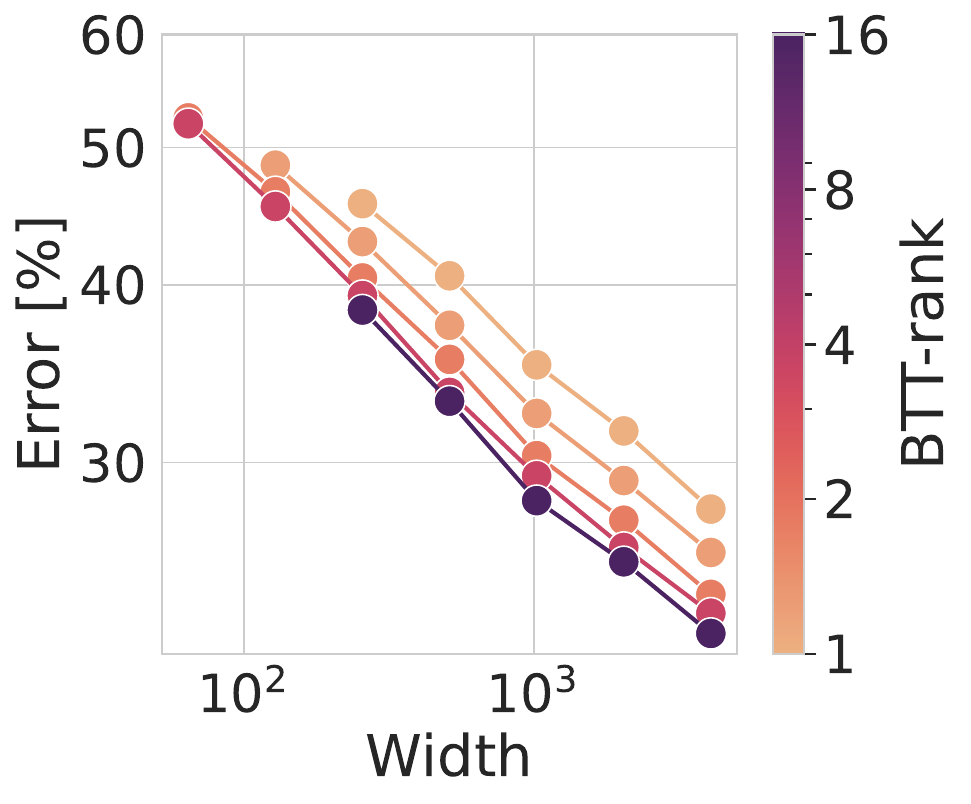}
    \label{fig:btt_rank_mem}
    }
    \subfloat[Monarch]{
    \includegraphics[width=0.5\linewidth]{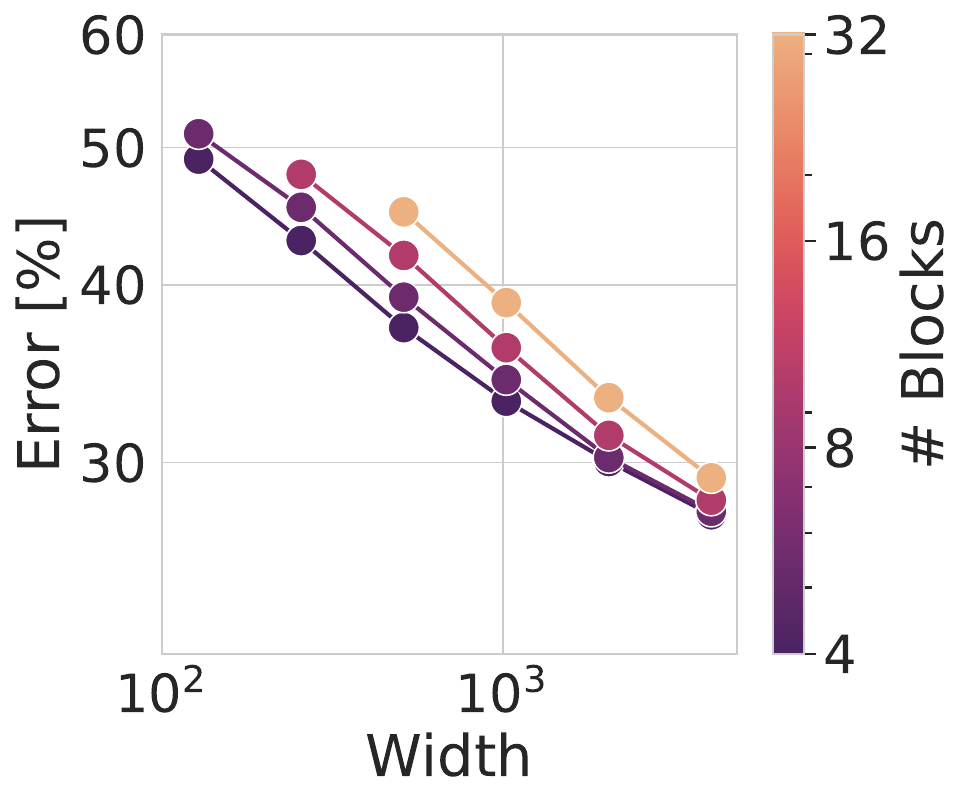}
    \label{fig:blocks_mem}
    }
   \caption{ \textbf{More compute per dimension is more memory-efficient on CIFAR-10.}
   (a) BTT with a higher rank achieves lower train error per unit width.
   (b) Monarch with fewer blocks achieves lower train error per unit width. A smaller width means less memory is required to store the activations. A lighter color indicates less compute per dimension.
   }
    \label{fig:memory}
    \vspace{-4mm}
\end{figure}

\noindent \textbf{Optimizing compute spent per dimension.} \quad Both BTT and Monarch have hyperparameters (BTT-rank $r$ and number of blocks $b$) that control how well they can approximate a dense matrix of the same dimension. We can scale up the compute $C$ in a structured layer by increasing either its dimension $d$ or its compute per dimension $\xi:=C/d$ (compute cost for an MVM normalized by $d$), which is controlled by these hyperparameters. From \Cref{tab:structures_memory_compute}, the compute per dimension is $d$ for dense, $2r\sqrt{d}$ for BTT (with 2 cores), and $2d/b$ for Monarch. To maximize performance as a function of $C,$ we need to optimally allocate it between the dimension $d$ of the layer and the compute spent per dimension $\xi$. In \Cref{fig:btt_rank_flops}, we show that while higher rank BTTs scale better than dense matrices on CIFAR-10, lower rank BTTs are more compute-efficient. Similarly, in \Cref{fig:blocks_flops}, Monarch matrices with more blocks and higher sparsity are more compute-efficient. These results illustrate that the optimal compute per dimension on this task is much smaller than $d$, and structured matrices beat dense matrices by making a favorable trade-off between dimension and compute per dimension.
In \Cref{app:more-cores}, we show that for BTT with $c \geq 3$ cores and different BTT-ranks, smaller ranks lead to better compute-efficiency, and using $c$ greater than 2 does not significantly improve compute efficiency on CIFAR-10, despite compute per dimension scaling as $\order{d^{1/c}}$.

The optimal way to scale $\xi$ with $d$ is likely non-trivial and task-dependent. The extremes are $\xi = d$ for a dense matrix and $\xi=0$ for the identity. The latter is clearly suboptimal, and neither is the former in light of our findings.

\noindent \textbf{Compute-memory trade-off.} \quad While lowering the compute per dimension can increase compute efficiency, it sacrifices memory efficiency if the memory cost is dominated by storing activations, such as when training with large batch sizes. In this case, the memory for storing activations scales at least as the layer width $d$. Since we can increase the expressivity of BTT and Monarch by increasing the rank or decreasing the number of blocks without increasing $d$, these hyperparameters enable us to trade off compute-efficiency with memory-efficiency, as demonstrated in \Cref{fig:memory}. While dense matrices are the least compute-efficient, they are the most efficient in terms of activation memory by packing the most parameters and compute into each dimension. The most compute-efficient yet memory-feasible structure will vary depending on the specific memory budget.

\section{Training Structured Transformers} \label{sec:trans}
We now apply structured layers to train larger transformer models for ImageNet classification and language modeling. We also introduce a technique required to prevent training divergence in these experiments.

\subsection{Stabilizing Training with Weight Normalization}
\begin{figure}[!t]
\centering
    \includegraphics[width=0.6\linewidth]{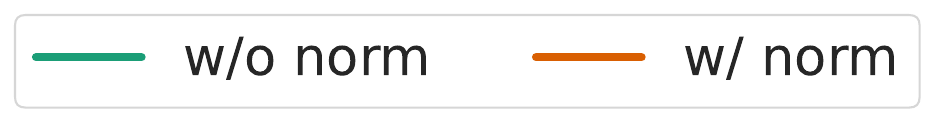}
    \\
    \subfloat[Activation growth]{
    \includegraphics[width=0.48\linewidth]{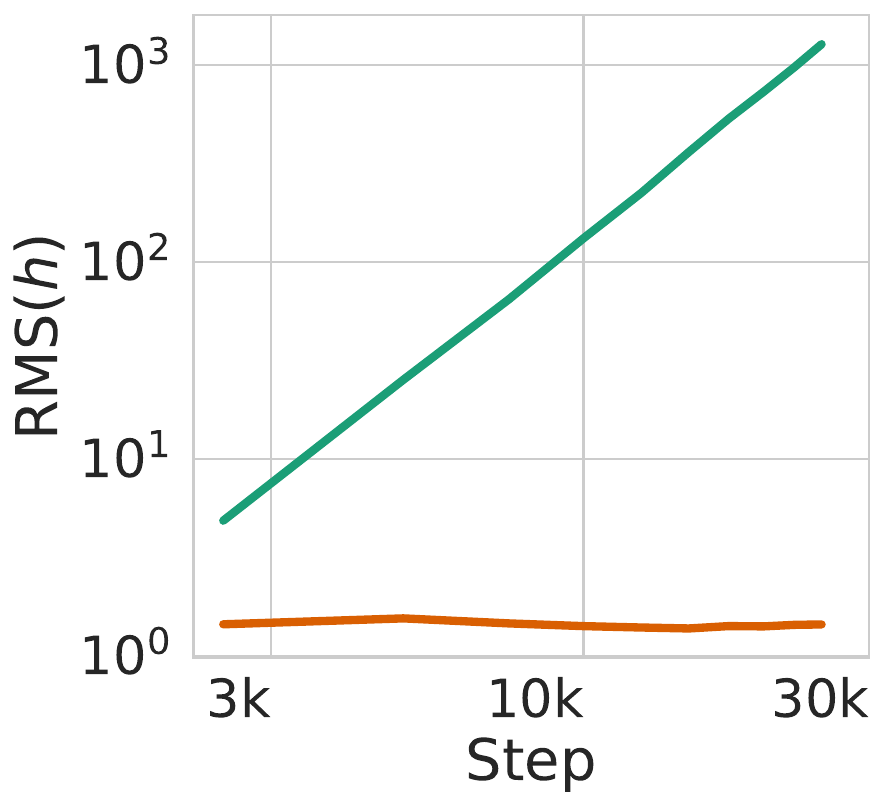}
    \label{fig:blowup}
    }
    \subfloat[Validation loss]{
    \includegraphics[width=0.48\linewidth]{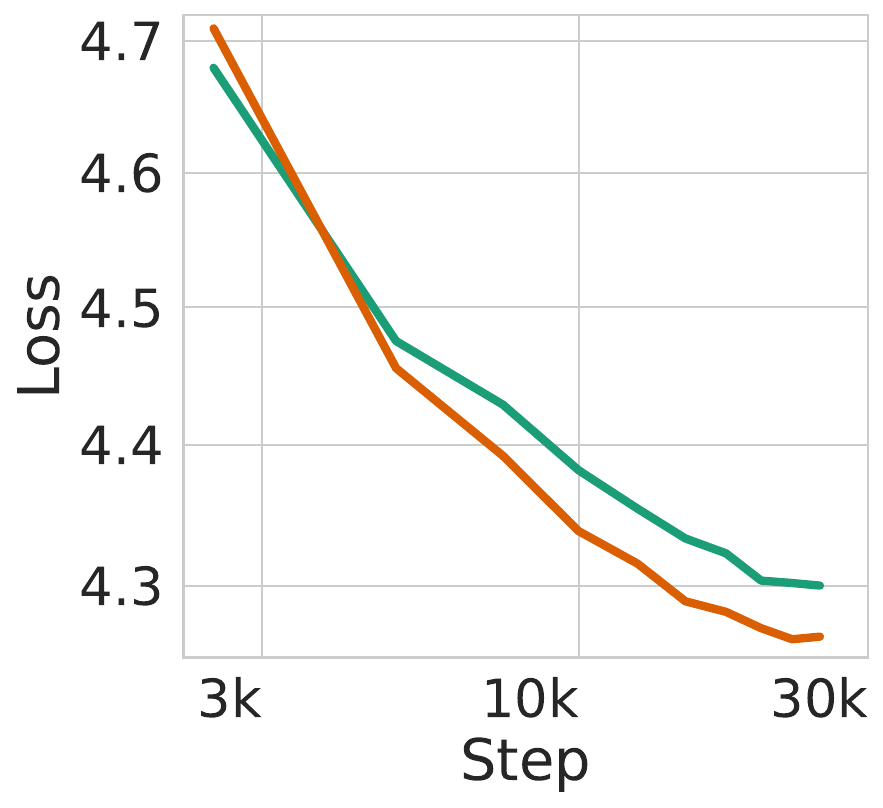}
    } \\
   \caption{
   \textbf{Weight normalization is necessary to stabilize GPT-2 training with BTT.} (a) RMS entry size of the final layer activations stabilizes around 1 with normalization but grows without bound otherwise. (b) Normalization improves validation loss.
   }
    \label{fig:btt_norm}
    \vspace{-5mm}
\end{figure}

When training on ImageNet and OpenWebText with BTT layers, we found the activations grow without bound slowly over time as illustrated in \Cref{fig:blowup} for GPT-2, which does not happen in the dense model.
We found we can eliminate this behavior without sacrificing expressivity through the following reparameterization:
\begin{align*}
    \bm{\tilde{M}} = \gamma_\bm{M} \min\qty(1, \frac{\sigma_\bm{M}}{\mathrm{RMS}(\bm{M})}) \bm{M},  
\end{align*}
where $\bm{M} \in \{\bm{L}, \bm{R}\}$. It normalizes the BTT cores $\bm{L}$ and $\bm{R}$ to have RMS entry sizes no larger than their initialization scales $\sigma_\bm{L}$ and $\sigma_\bm{R}$, and scaling them by learnable scalars $\gamma_\bm{L}$ and $\gamma_\bm{R}$ to allow the singular values to grow in size if needed, similar to what is proposed in \citet{salimans2016weight}.  In \Cref{fig:btt_norm}, we show a 12-layer GPT-2 model with $d = 128$ using BTT layers trained on OpenWebText with or without normalization. Weight normalization eliminates the unbounded growth of the activations before the last layer normalization, which will eventually lead to NaN. Weight normalization also improves validation loss, which is in contrast to alternatives such as lowering the learning rate and increasing weight decay which we found to only reduce the rate of growth at the cost of worse performance.

\begin{figure}[!t]
\centering
    \includegraphics[width=0.8\linewidth]{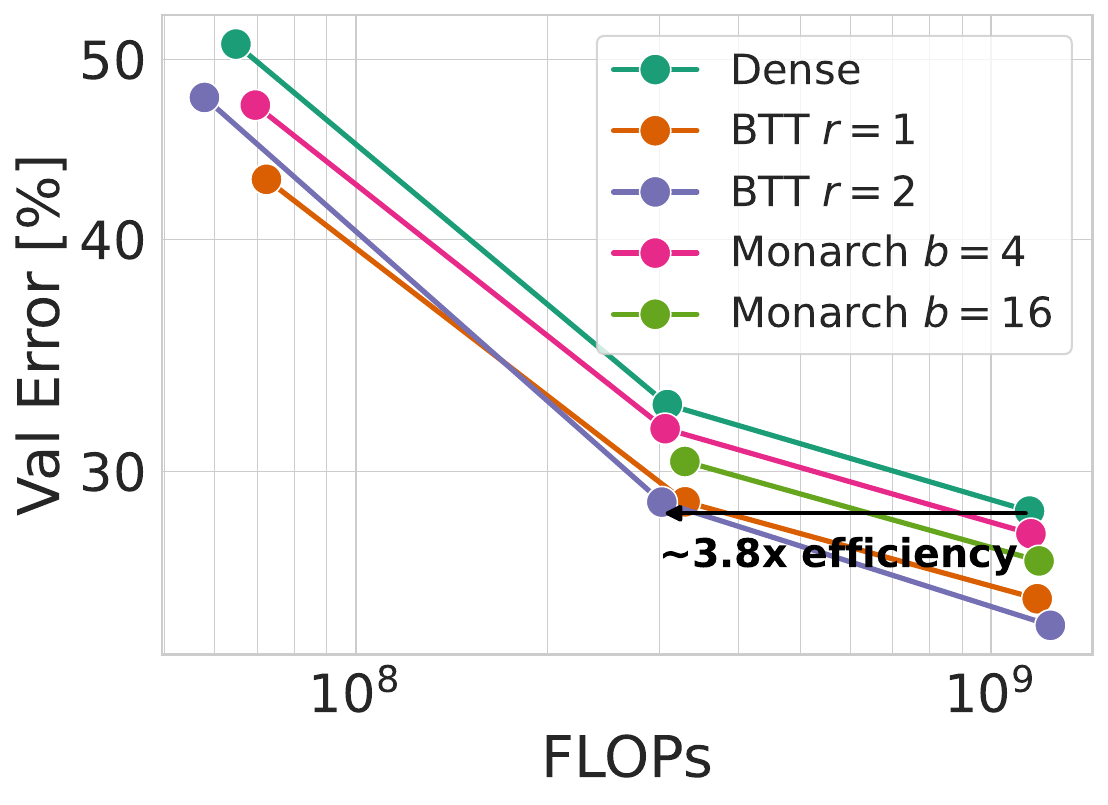}
   \caption{
   \textbf{ViTs trained on ImageNet with structured layers are more compute-efficient}. We use ViTs with patch size 32 trained for 300 epochs. BTT reaches the same performance of a dense ViT-S/32 with up to $3.8\times$ fewer FLOPs. 
   }
    \label{fig:imagenet}
\end{figure}

\subsection{ViT on ImageNet}
We train ViTs with patch size $32$ on ImageNet for 300 epochs. We provide full experimental details in \Cref{app:trans}.
In \Cref{fig:imagenet}, we find both BTT with rank $r \in \{1, 2\}$ and Monarch with $b \in \{4, 16\}$ blocks outperform dense for the same amount of compute for training ViTs on ImageNet. BTT reaches the same performance of a dense ViT-S/32 (the larger dense model shown) with up to $3.8\times$ fewer FLOPs. We find Monarch with 16 blocks is more compute-efficient than with 4 blocks, the original version used in \citet{dao2022monarch} and in the Monarch Mixer architecture \citep{fu2023mixer}, consistent with our finding on CIFAR-10 that less compute per dimension is more compute-efficient.

\begin{figure}[!t]
\centering
    \subfloat[All compute]{
    \includegraphics[width=0.48\linewidth]{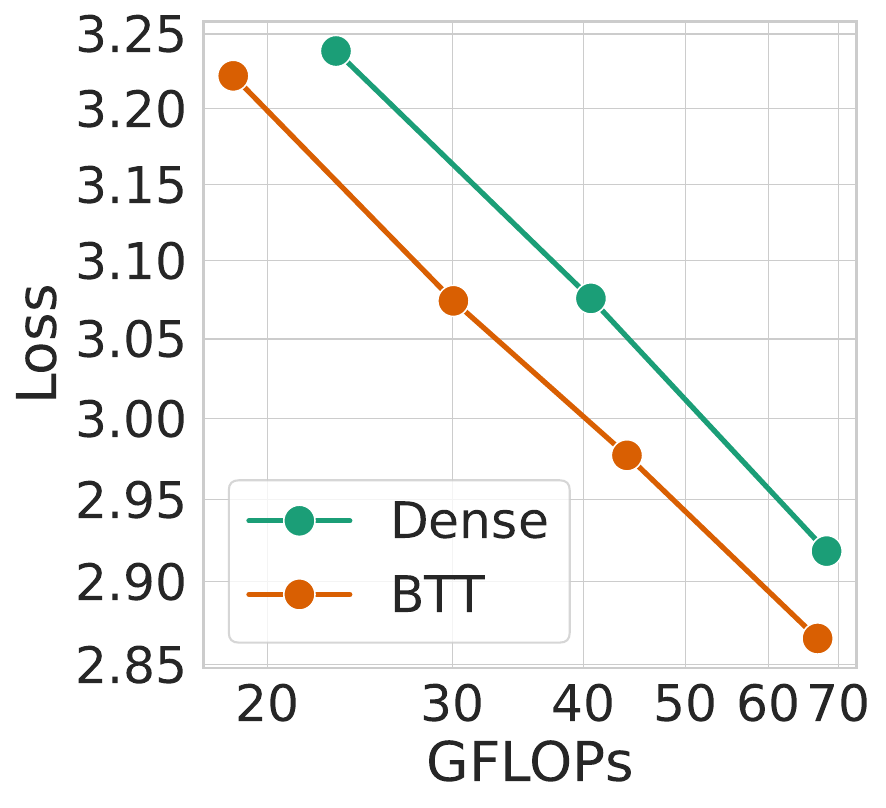}
    \label{fig:gpt_all_compute}
    }
    \subfloat[Non-embedding compute]{
    \includegraphics[width=0.48\linewidth]{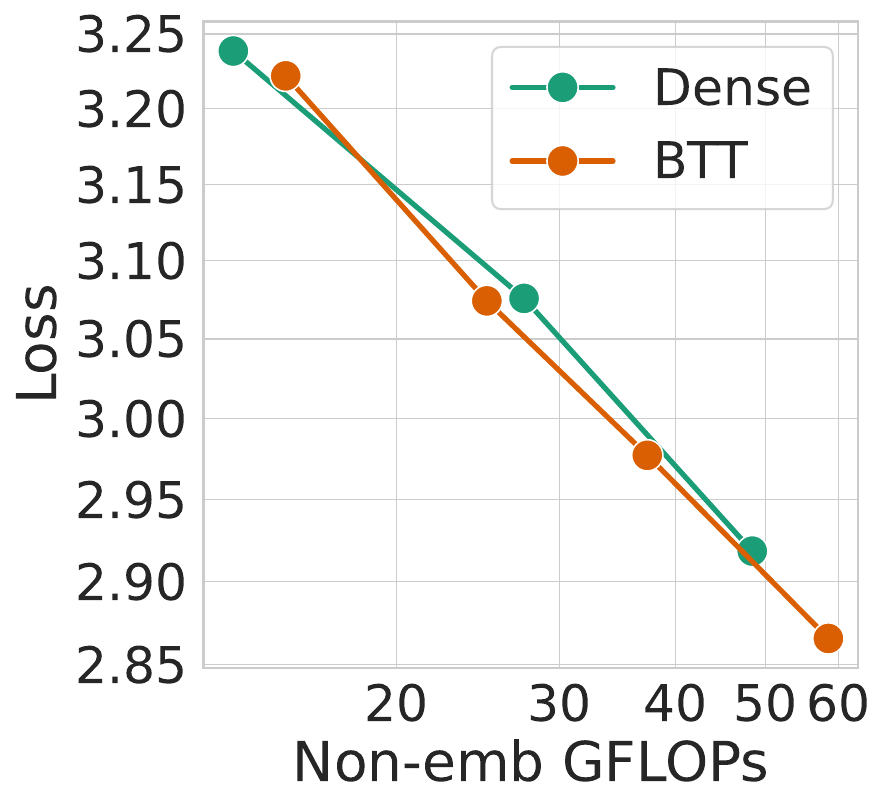}
    \label{fig:gpt_nonemb_compute}
    }
   \caption{
   \textbf{GPT-2 with all BTT layers is more compute-efficient}. (a) When including language modeling head compute, BTT is more efficient than dense. (b) When excluding language modeling head compute, BTT and dense perform similarly. 
   }
    \label{fig:gpt2}
    \vspace{-2mm}
\end{figure}

\subsection{GPT-2 on OpenWebText}
We train GPT-2 models on OpenWebText for $600,000$ steps with a batch size of $245,760$ tokens at a sequence length of $512. $ We provide full experimental details in \Cref{app:trans}. We replace all linear layers, including the language modeling head, which accounts for a significant fraction of the compute, with BTT layers.
In \Cref{fig:gpt_all_compute}, we show the resulting GPT-2 model with BTT layers outperforms the original dense GPT-2 as a function of compute. However, in \Cref{fig:gpt_nonemb_compute}, we find they perform similarly when controlling for non-embedding compute, which excludes the compute spent in the language modeling head \citep{kaplan2020scaling}. While the improvement is significant, \Cref{fig:gpt_nonemb_compute} suggests that the improvement primarily comes from reducing the compute spent in the language modeling head and may therefore diminish at larger scales where the fraction of compute spent in the language modeling head becomes negligible.

\section{Discussion} \label{sec:discussion} 

The exponential growth in the computational cost of training foundation models in recent years has made the development of more compute-efficient architectures and training procedures a critical area of research. While structured matrices have traditionally been used in machine learning to approximate dense matrices or encode constraints such as equivariance, our work shows their promise in serving as general-purpose linear layers, a universal compute bottleneck in current foundation models, while offering improved compute efficiency relative to dense matrices.

Our work uncovers several key insights in designing more compute-efficient linear layers with structured matrices:
\begin{itemize}
    \item \textit{Careful optimization is crucial:} structure-aware learning rates based on \mup are essential to realize the performance benefits of structured matrices.
    \item \textit{Better scaling laws than dense are possible:} structured matrices can sometimes exponentially outperform dense matrices as we increase compute.
    \item \textit{Relaxing parameter sharing produces compute-efficient and general-purpose structures:} By learning more parameters with the same compute, Monarch and BTT can provide better performance as general linear layers than the parameter-sharing Kronecker product and Tensor-Train structures.
    \item \textit{Compute per dimension is an impactful yet neglected hyperparameter:} dense matrices consume the most compute per dimension, but they can underperform structured matrices that trade less compute per dimension for more dimensions, resulting in wider models.
\end{itemize}

Extending our evaluation to larger-scale models and datasets, studying the compute-optimal scaling laws, and developing a theoretical understanding of when and why structured matrices can improve scaling laws based on data and model characteristics are exciting directions for future work.

\section*{Acknowledgements} We thank Sanae Lotfi, Alan Amin, and Bayan Bruss for helpful discussions, and Christopher Ferri
 for HPC assistance. This work is supported by NSF CAREER IIS-2145492,
NSF CDS\&E-MSS 2134216, NSF HDR-2118310, BigHat Biosciences, Capital One, and an Amazon Research Award.

\clearpage

\section*{Impact Statement}
This work aims to improve the performance of MLPs and transformers per unit of compute.  Making neural networks more efficient has the potential to reduce energy consumption of training and inference, and more efficient neural networks can also make deep learning accessible where compute resources are scarce.  However, we caution that the matrix structures we use should be tested in new domains, at new architectural scales, and within new architectures, to ensure that our results extrapolate for a practitioner's specific individual needs.

\bibliography{refs}
\bibliographystyle{icml2024}

\newpage
\appendix
\onecolumn

\section{Related Work}
\paragraph{Compute-Efficient Alternatives to Dense Layers.}
Finding more compute-efficient alternatives to dense layers during training is an under-explored research topic. Convolutional networks and other equivariant models using structured matrices only offer an advantage in specific domains where the assumed symmetries exist \citep{lecun1998gradient, finzi2020generalizing}. Approaches such as pruning and quantization \citep{han2016deep, molchanov2016prun, liu2017slim, frankle2018ticket, mishra2021accelerating} mainly target reducing the inference cost after a model has been trained. Similarly, \citet{lee2023differentiable} introduce a differentiable approach to learn a sparse structure that contain sums of low-rank blocks, but the learned structure can only be made sparse after training. Efficient fine-tuning methods leveraging structured matrices, such as LoRA \citep{hu2021lora}, only apply in the fine-tuning stage. Recent works have used low-rank structures to reduce the memory usage of training and accelerate the backward pass, but they still use dense matrices in the forward pass \citep{zhao2024galore,lialin2023relora}. While Tensor-Train decomposition can improve parameter efficiency of neural networks \citep{chekalina2023gpttrans,novikov2015tt-nets}, they have not been shown to improve their compute efficiency. 

The recently proposed Monarch matrices \citep{dao2022monarch} are a notable exception, which enable faster training of certain vision and language transformers by training with Monarch matrices for all or most of the training steps followed by only a small amount of dense training.

\paragraph{Initialization and Learning Rate for Structured Layers.}
The most popular initialization strategies such as Xavier \citep{glorot2010understanding}, Kaiming \citep{he2015delving}, and Lecun \citep{lecun2002efficient} initializations set the initialization scales of the dense matrices so that the forward or backward pass is variance preserving at initialization. \citet{pan2022unified} extended this analysis to tonsorial convolutional networks where the kernels are structured. In addition to considering only a subset of possible structures (dense and tensorial convolution), these strategies are not optimal because they only consider the initialization and not the training dynamics, as shown by \mup \citep{yang2021v}. Specifically, \mup uses an asymptotically smaller initialization variance compared to these methods when a layer's input dimension is asymptotically larger than its output dimension, such as the last layer.

To the best of our knowledge, there is no prior work that investigates how to scale the learning rate for general structured linear layers. Prior works using Tensor-Train Decomposition \citep{chekalina2023gpttrans}, low-rank matrices \citep{lialin2023relora}, and Monarch matrices \citep{dao2022monarch} to replace dense layers simply used global learning rates for all parameters and do not specify how they should be scaled as a function of width. The concurrent work LoRA+ \citep{hayou2024lora+} studies the special case for low-rank matrices of the form $\bm{W} = \bm{U}\bm{V}, \bm{U} \in \R^{d\times r}, \bm{V} \in \R^{r\times d}, r \ll d,$ and proposes that $\bm{U}$ should have a higher learning rate compared to $\bm{V}$, consistent with the more general analysis we present in this work that also applies to other structured matrices.

\section{Runtime Comparisons} \label{app:runtime}

\begin{figure*}[!h]
\centering
    \subfloat[MVM Runtime vs FLOPs]{
    \includegraphics[height=0.25\linewidth]{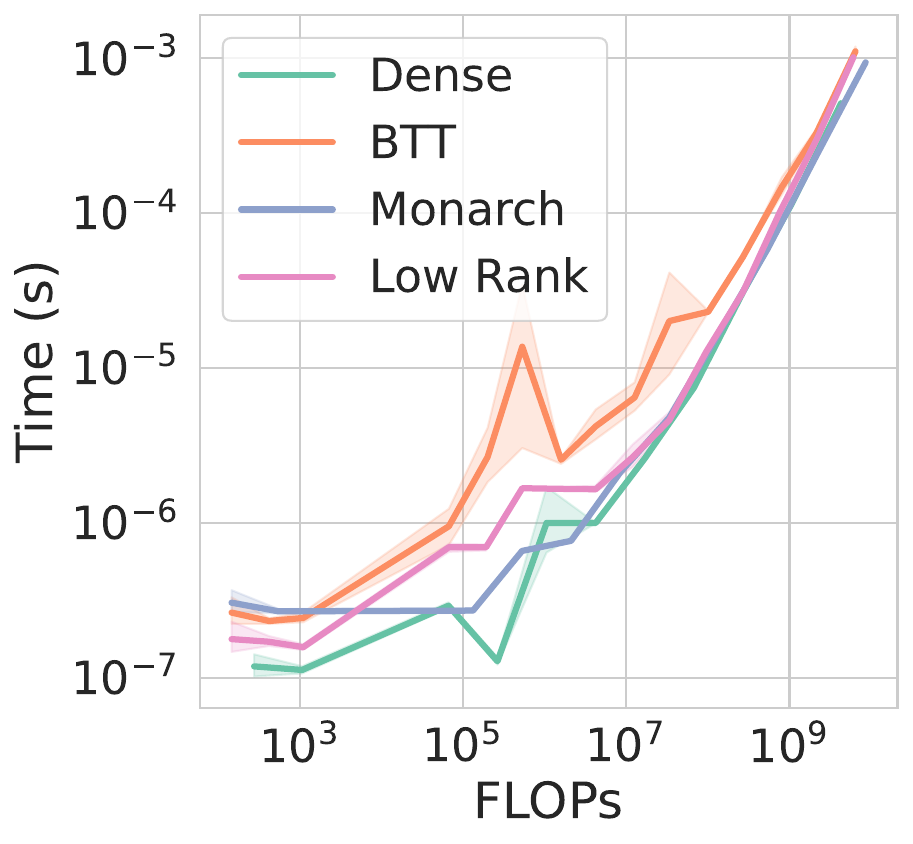}
    \label{fig:mvm_time}
    }
    \subfloat[CIFAR-100 Train Error vs FLOPs]{
    \includegraphics[height=0.25\linewidth]{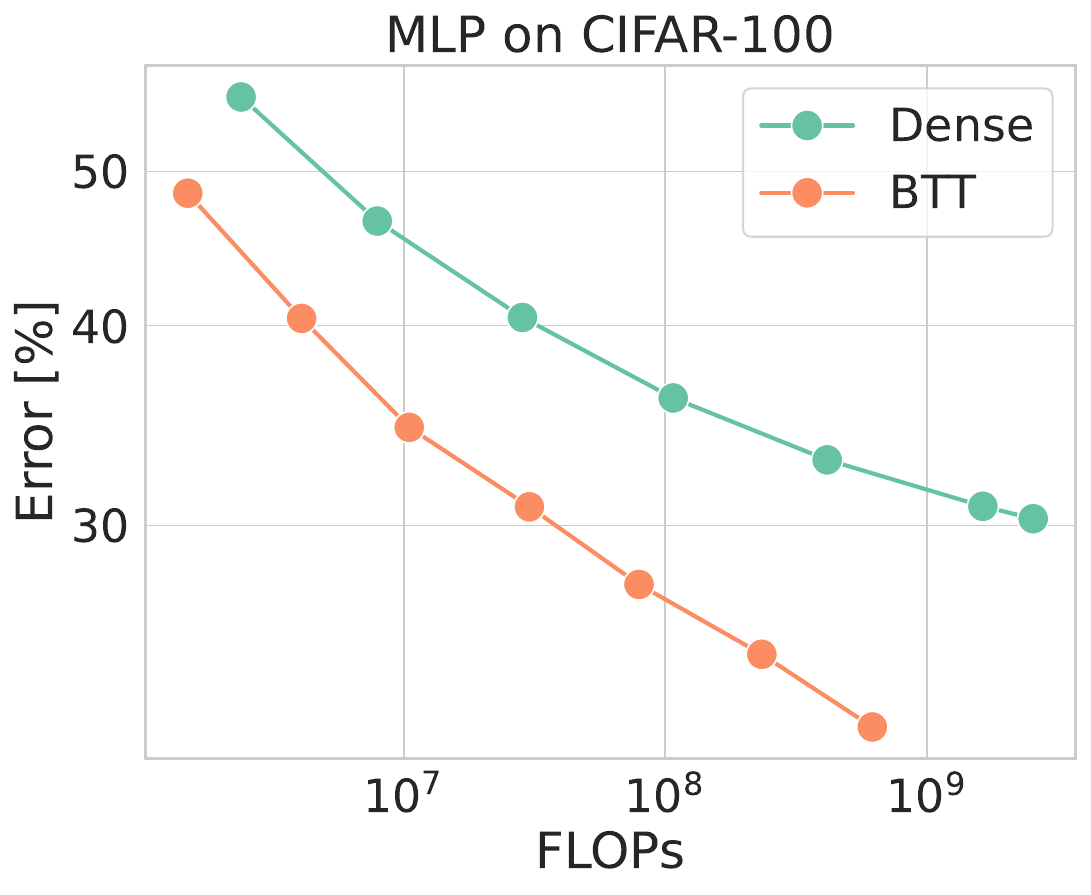}
    \label{fig:c10-flops}
    }
    \subfloat[CIFAR-100 Train Error vs Runtime]{
    \includegraphics[height=0.25\linewidth]{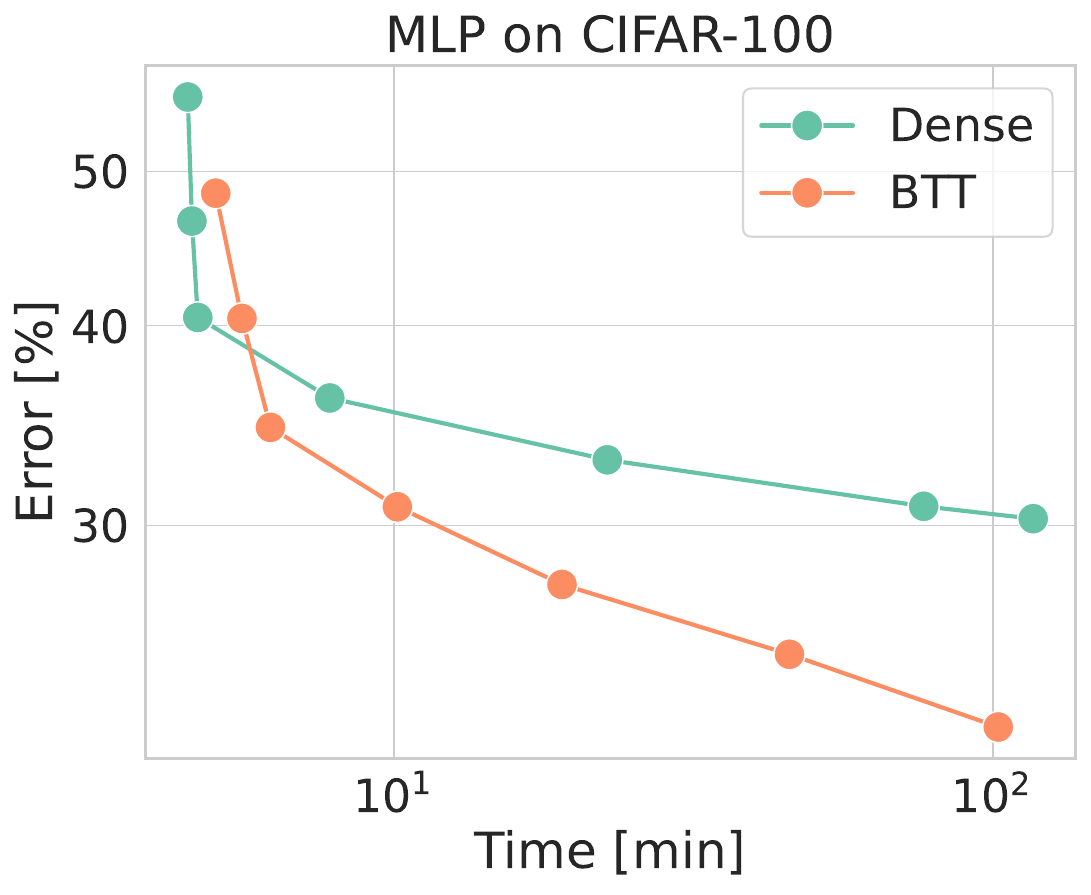}
    }
   \caption{
     \textbf{At large scales, runtime and FLOPs are equivalent for the structures we consider.} We omit Kronecker and TT in (a) because they are special cases of Monarch and BTT.
     }
    \label{fig:c10-time}
    \vspace{-4mm}
\end{figure*}

All structures in this work use the same dense matrix multiplication primitive on the GPU, so FLOPs are proportional to their runtimes for large matrix sizes. Only below a certain scale do runtimes vary noticeably between structures as a function of FLOPs. We verify this on an Nvidia A100 GPU in \Cref{fig:mvm_time}, showing the time for matrix-vector multiplication for different structures vs. FLOPs. For small matrices, runtimes vary between structures and don't reflect FLOPs due to inefficient tensor core utilization. For large matrices, runtimes converge to the same function in FLOPs. Optimizing structured matrix implementations can reduce their runtime overhead and will be essential to realizing the practical benefits of these structures.

Measuring FLOPs allows incorporating results from smaller experiments without letting the runtime inefficiencies at small scale obscure the scaling laws. \Cref{fig:c10-flops}  and \Cref{fig:c10-time} compare BTT with dense MLPs on CIFAR-100 in FLOPs and runtimes on an Nvidia A100. Below $\sim10^7$ FLOPs, increasing FLOPs barely changes runtimes for dense and BTT, obscuring the scaling laws. BTT underperforms dense when controlling for runtime by incurring longer runtime per FLOP at this scale. However, as compute increases, scaling laws in FLOPs translate to scaling laws in runtimes, with BTT outperforming dense significantly.

\section{General Expression for Tensor-Train and Block Tensor-Train} \label{app:general}
Here we describe the general expression for Tensor-Train and Block Tensor-Train, with an arbitrary number of cores and ranks. To make the expression more intuitive, we will use superscripts for output indices and subscripts subscripts for input indices. Rank indices appear once as a superscript when first introduced and once as a subscript when summed away.

\begin{table}[!t]
    \centering
    \begin{tabular}{l|c}
       Structure & Learning rate multiplier $\kappa$ \\
      \hline
      Low-Rank $\bm{U} \bm{V}$ & $\kappa_\bm{U} = d/2r, \kappa_\bm{V} = 1/2$ \\
      \hline
      Kronecker $\bm{L} \otimes \bm{R}$ & $\kappa_\bm{L} = \sqrt{d}/2, \kappa_\bm{R} = \sqrt{d}/2$ \\
      \hline
      Monarch $\bm{P} \bm{L} \bm{P}^{\top} \bm{R}$ & $\kappa_\bm{L} = b/2, \kappa_\bm{R} = b/2$ \\
      \hline
      TT$(\bm{L},\bm{R})$ & $\kappa_\bm{L} = \sqrt{d}/2r, \kappa_\bm{R} = \sqrt{d}/2$ \\
      \hline
      BTT$(\bm{L},\bm{R})$ & $\kappa_\bm{L} = \sqrt{d}/2r, \kappa_\bm{R} = \sqrt{d}/2$ \\
    \end{tabular}
    \caption{
      \textbf{Learning rate multipliers for structured matrices.} We show the Adam learning rate multiplier $\kappa$ we use for each parameter tensor of the structure when transferring the learning rate from a dense layer of the same width $d.$ $r$ refers to the rank in low rank, TT, and BTT, while $b$ refers to the number of blocks in Monarch. 
    }
    \label{tab:structures_mup}
\end{table}

\noindent \textbf{Tensor-Train.} \quad
Tensor-Train (TT) decomposition of a $\dout \times \din$ matrix $\bm{W}$ is defined by a set of $c$ cores $\bm{G}_t \in \mathbb{R}^{r_{t-1} \times  m_{t} \times  n_{t} \times  r_{t}}$
for $t=1, \dots, c,$ where $c \geq 2, \dout = \prod_{t}^{} m_{t}, \din = \prod_{t}^{} n_{t},$ $r_{0} = r_{c} = 1$ and $\{r_t\}_{t=1}^{c}$ being free integer hyperparameters. These cores specify the elements of an
$n_1 \times \ldots \times n_t \times m_1 \times \ldots \times m_t$ tensor $\bm{T}$ via

\begin{equation}
    T^{i_1, \ldots, i_c}_{j_1, \ldots, j_c} = \sum_{\alpha_1, \ldots, \alpha_{t+1}} \prod_{t=1}^{c} (G_t)^{\alpha_{t-1}, i_t}_{ j_t, \alpha_t}.
\end{equation}

Identifying elements of $\bm{T}$ with elments of a $\dout \times \din$ matrix $\bm{W}$, the efficient matrix-vector multiply against $\bm{W}$ does not involve materializing $\bm{W}$ but is simply given by a sequence of contractions against each core $\bm{G}_t$ from $t=c$ to $t=1:$

\begin{equation} \label{eq:full}
    \begin{split}
      (z_{t-1})^{\alpha_{t-1}, j_1, \ldots, j_{t-1}, i_t, \ldots, i_c} = \sum_{\alpha_t=1}^{r_t}  \sum_{j_t=1}^{n_t} (G_t)^{\alpha_{t-1}, i_t}_{j_t, \alpha_t} (z_t)^{\alpha_t, j_1, \ldots, j_t, i_{t+1}, \ldots, i_c},
    \end{split}
\end{equation}

where the initial $\bm{z}_c$ is obtained by reshaping the input $\bm{x}$ into an $n_c \times n_{c-1} \ldots \times n_1 \times 1$ tensor and the final $\bm{z}_0$ is flattened into an output vector. Suppose, for convenience, $\din=\dout=d,$ $n_t = m_t = d^{1/c}$ for all $t,$ and $r_t = r$ for all $t \notin \{0, c\},$ then TT has $P=(2r + (c-2) r^2) d^{2/c}$ parameters, and an MVM costs $C=(2r + (c-2) r^2) d^{1+c^{-1}}$ FLOPs. Note we have $C = P d^{1 - c^{-1}},$ showing each parameter is used for $d^{1 - c^{-1}} \geq \sqrt{d}$ times.

\noindent \textbf{Block Tensor-Train.} \quad
Block Tensor-Train (BTT) is defined simply by appending \tc{additional axes} to each core in TT via the substitution
\begin{equation}
    (G_t)^{\alpha_{t-1}, i_t}_{j_t, \alpha_t} \to (G_t)^{\alpha_{t-1}, i_t, \tc{i_{t+1}, \ldots, i_c}}_{\tc{j_1, \ldots, j_{t-1}}, j_t, \alpha_{t}}.
\end{equation}

As before, multiplying the cores and summing out the rank axes, we have
\begin{equation}
    T^{i_1, \ldots, i_c}_{j_1, \ldots, j_c} = \sum_{\alpha_1, \ldots, \alpha_{t+1}} \prod_{t=1}^{c} (G_t)^{\alpha_{t-1}, i_t, \tc{i_{t+1}, \ldots, i_c}}_{\tc{j_1, \ldots, j_{t-1}}, j_t, \alpha_{t}}.
\end{equation}

Efficient multiplication with the corresponding matrix is now given by

\begin{equation}
    \begin{split}
      (z_{t-1})^{\alpha_{t-1}, j_1, \ldots, j_{t-1}, i_t, \ldots, i_c} = \sum_{\alpha_t=1}^{r_t}  \sum_{j_t=1}^{n_t} (G_t)^{\alpha_{t-1}, i_t, \tc{i_{t+1}, \ldots, i_c}}_{\tc{j_1, \ldots, j_{t-1}}, j_t, \alpha_{t}} (z_t)^{\alpha_t, j_1, \ldots, j_t, i_{t+1}, \ldots, i_c},
    \end{split}
\end{equation}

which costs the same FLOPs as for TT, while admitting more learnable parameters. Again we do not need to materialize $\bm{T}.$ Suppose, for convenience, $\din=\dout=d,$ $n_t = m_t = d^{1/c}$ for all $t,$ and $r_t = r$ for all $t \notin \{0, c\},$ then BTT has $P=(2r + (c-2) r^2) d^{1+c^{-1}}$ parameters, equal in number to the FLOPs for an MVM $C=(2r + (c-2) r^2) d^{1+c^{-1}}$. Thus, for the same amount of compute, BTT can learn a factor of $d^{1 - c^{-1}} \geq \sqrt{d}$ more parameters than TT.

\section{Expressivity of Block Tensor-Train}
We start by providing an algorithm to approximate any existing dense matrix $\bm{A}$ with a BTT. The algorithm will then illustrate the expressivity of the BTT structure as a function of $c$ and $\{r_t\}_{t=1}^{c}$. For simplicity, we will assume $\bm{A} \in \R^{d \times d},$ and the cores will be square, having size $d^{1/c}$ in each dimension, except for the rank dimension. Generalization to non-square $\bm{A}$ and non-square cores is straigtforward.

\noindent \textbf{Projection onto Block Tensor-Train with $c=2$.} \quad
In the case where $c=2,$ we prove a closed-form expression for projecting an arbitrary dense matrix $\bm{A}$ to the closest rank-$r$ (there is only one rank parameter so we omit the subscript) BTT $\bm{B}$ that minimizes the squared Frobenius norm $\norm{\bm{A} - \bm{B}}^2_F.$ Writing $\bm{A}$ and $\bm{B}$ as $\sqrt{d} \times \sqrt{d} \times \sqrt{d} \times \sqrt{d}$ tensors with $B^{ii'}_{jj'} = \sum_{\alpha=1}^{r} L^{ii'}_{j\alpha} R^{\alpha i'}_{j j'},$ we have
\begin{align}
    & \norm{\bm{A} - \bm{B}}^2_F  \\
    =& \sum_{ii'jj'} \qty(A^{ii'}_{jj'} - \sum_{\alpha=1}^{r} L^{ii'}_{j\alpha} R^{\alpha i'}_{j j'})^2 \\
    =& \sum_{\tc{i'j}} \sum_{ij'} \qty(A^{i\tc{i'}}_{\tc{j}j'} - \sum_{\alpha=1}^{r} L^{i\tc{i'}}_{\tc{j}\alpha} R^{\alpha \tc{i'}}_{\tc{j} j'})^2 \\
    =& \sum_{\tc{i'j}}  \norm{\bm{A}^{(\tc{i'j})} - \sum_{\alpha=1}^{r} \bm{\ell}^{(\tc{i'j})}_\alpha \bm{r}^{(\tc{i'j})\top}_\alpha}^2_F, \\
\end{align}
where we have decomposed the minimization problem into multiple independent minimization problems: for each $i',j,$ we wish to find the best rank-$r$ approximation $\sum_{\alpha=1}^{r} \bm{\ell}^{(i'j)}_\alpha \bm{r}^{(i'j)\top}_\alpha$ to the matrix $\bm{A}^{(i'j)} \in \R^{\sqrt{d} \times \sqrt{d}}.$ Thus, we obtain an optimal solution by finding these best rank-$r$ approximation (e.g. via SVD) for each $\bm{A}^{(i'j)},$   and reassembling the vectors $\bm{\ell}^{(i'j)}_\alpha$ and $\bm{r}^{(i'j)}_\alpha$ into the tensors $\bm{L}$ and $\bm{R}.$ This result is a straightforward generalization of the algorithm for projection onto Monarch matrices \citep{dao2022monarch}, which deals with the case where $r=1.$

\noindent \textbf{Generalization to $c > 2$.} \quad For convenience, let's relabel $\bm{L}$ found in the previous algorithm as $\bm{\tilde{L}},$ and the rank $r$ as $r_2.$ Having found $\bm{\tilde{L}}$ and $\bm{R},$ we can recursively apply the above algorithm on $\bm{\tilde{L}}$ to find its optimal 2-core rank-$r_1$ BTT approximation, with cores $\bm{L}$ and $\bm{M}$. Together, $\bm{L}, \bm{M},$ and $\bm{R}$ parameterize a 3-core BTT approximation with ranks $r_1$ and $r_2.$ Similar to the recursive TT-SVD algorithm \citep{oseledets2011tt}, the found solution will not necessarily be optimal for $c > 2$ due to its greediness.

It is sufficient to illustrate this algorithm in detail for $c=3.$ Reshaping $\bm{A}$ into a tensor $A^{i_1 i_2 i_3}_{j_1 j_2 j_3} \in \R^{d^{1/3}\times\ldots\times d^{1/3}},$ we wish to find $B^{i_1 i_2 i_3}_{j_1 j_2 j_3} = \sum_{\alpha=1}^{r} \sum_{\beta=1}^{r} L^{i_1 i_2 i_3}_{j_1 \beta} M^{\beta i_2 i_3}_{j_1 j_2 \alpha} R^{\alpha i_3}_{j_1 j_2 j_3}$ that approximates $\bm{A}.$ We first group $i_1,i_2$ as a single index $(i_1 i_2)$ and $j_1,j_2$ as a single index $(j_1 j_2),$ and then apply the previous algorithm for the 2-core case to find $\tilde{\bm{L}}, \bm{R}$ that minimizes
\begin{equation}
    \sum_{(i_1 i_2) i_3 (j_1 j_2) j_3} \qty(A^{(i_1 i_2) i_3}_{(j_1 j_2) j_3} - \sum_{\alpha=1}^{r_2} \tilde{L}^{(i_1 i_2) i_3}_{(j_1 j_2) \alpha} R^{\alpha i_3}_{(j_1 j_2) j_3})^2,
\end{equation}
forming the best following best rank-$r_2$ 2-core approximation:
\begin{equation}
    \label{eq:2core}
    A^{(i_1 i_2) i_3}_{(j_1 j_2) j_3} \approx \sum_{\alpha=1}^{r_2} \tilde{L}^{(i_1 i_2) i_3}_{(j_1 j_2) \alpha} R^{\alpha i_3}_{(j_1 j_2) j_3}.
\end{equation}
Setting $r_2 = \min(\#(i_1 i_2), \# j_3) = \sqrt{d}$ will lead to an exact decomposition, where $\# \chi$ denotes the length of the range of the index $\chi$.
Then we un-group the indicies to the obtain $\tilde{L}^{i_1 i_2 i_3}_{j_1 j_2 \alpha}, R^{\alpha i_3}_{j_1 j_2 j_3}.$ Now grouping $i_2 i_3$ and $j_2 \alpha$ as single indices, we apply the previous algorithm again to find the best rank-$r_1$ 2-core BTT approximation to $\tilde{\bm{L}}$ yielding the tensors $\bm{L}, \bm{M}$ that minimize
\begin{equation}
    \sum_{i_1 (i_2 i_3) j_1 (j_2 \alpha)} \qty(\tilde{L}^{i_1 (i_2 i_3)}_{j_1 (j_2 \alpha)} - \sum_{\beta=1}^{r_1} L^{i_1 (i_2 i_3)}_{j_1 \beta} M^{\beta (i_2 i_3)}_{j_1 (j_2\alpha)})^2. \\
\end{equation}

Setting $r_1 = \min(\# i_1, \# (j_2 \alpha)) = \sqrt{d}$ will again lead to an exact decomposition,
Now replacing $\tilde{L}^{i_{12} i_3}_{j_{12} \alpha}$ in \Cref{eq:2core} by its approximation $\sum_{\beta=1}^{r_1} L^{i_1 i_2 i_3}_{j_1 \beta} M^{\beta i_2 i_3}_{j_1 j_2\alpha},$ we have found the 3-core BTT approximation to $\bm{A}$ with ranks $(r_1, r_2):$
\begin{equation}
    A^{i_1 i_2 i_3}_{j_1 j_2 j_3} \approx B^{i_1 i_2 i_3}_{j_1 j_2 j_3} = \sum_{\beta=1}^{r_1} \sum_{\alpha=1}^{r_2} L^{i_1 i_2 i_3}_{j_1 \beta} M^{\beta i_2 i_3}_{j_1 j_2 \alpha} R^{\alpha i_3}_{j_1 j_2 j_3}.
\end{equation}

\noindent \textbf{Quantifying the expressivity of BTT.} \quad
By applying the above recursive algorithm and always choosing a high enough rank so that the decomposition is exact at each step, we prove that a $c$-core BTT with sufficiently large ranks $\{r_t\}_{t=1}^{c}$ can represent any $d\times d$ dense matrix exactly. Moreover, the general expression for an upper-bound on $r_t$ to ensure exact decomposition can be deduced as $r_t \leq \min(\# i_1 \times \ldots \times \# i_t, \# j_{t+1} \times r_{t+1}) \leq d^{\min(t, c-t)/c}:$ i.e. $r_1 \leq d^{1/c}, r_2 \leq d^{2/c}, \ldots, r_{c/2} \leq \sqrt{d}, \ldots, r_{c-1} \leq d^{2/c}, r_c \leq d^{1/c}.$
By contrast, TT has a worse bound of $r_1 \leq d^{2/c}, r_2 \leq d^{4/c}, \ldots, r_{c/2} \leq d, \ldots, r_{c-1} \leq d^{4/c}, r_c \leq d^{2/c}$ \citep{oseledets2011tt}.

A practical takeaway is that we can monotonically improve the expressivity of BTT by increasing $r_t$ until the bound is reached, and we should never use ranks larger than the bound since it creates unnecessary redundancy in the parameterization.

\section{Scaling Laws Experiment Details} \label{app:exp-details}
We provide code for reproducing our experiments \href{https://github.com/shikaiqiu/compute-better-spent}{\underline{here}}.
\subsection{Model architectures} \label{app:arch}
\noindent \textbf{MLP.} \quad
Following \citet{bachmann2023scaling}, we use MLPs consisting of residual blocks of the form
\begin{equation}
      \bm{h}_{\ell+1}
      =
      \bm{h}_{\ell} + \bm{W}_{\ell}^{(2)}g\qty(\bm{W}_{\ell}^{(1)} \text{LN}\left(\bm{h}_{\ell}\right)), \quad \bm{W}_{\ell}^{(1)} \in \R^{4d \times d}, \quad \bm{W}_{\ell}^{(2)} \in \R^{d \times 4d},
\end{equation}
where $g\left(\cdot\right)$ denotes the GELU activation \citep{hendrycks2016gelu}
and $\text{LN}\left(\cdot\right)$ stands for layer normalization \citep{ba2016ln}. In addition, there is an input embedding layer and a classification layer. We refer to $d$ as the width of the model. We use models with $3$ residual blocks and scale them up by increasing $d.$

\noindent \textbf{ViT.} \quad
We use standard ViTs \citep{dosovitskiy2022vit}, but with $1/d-$scaled rather $1/\sqrt{d}-$scaled attention as prescribed by \mup \cite{yang2021v} and Query-Key Normalization \citep{henry2020query, wortsman2023small} for improved stability. We refer to the embedding dimension, commonly denoted $d_\mathrm{model}$, as the width $d$ of the model. We use models with $3$ transformer blocks and scale them up by increasing $d.$

\subsection{Hyperparameters}
\noindent \textbf{Training hyperparameters.} \quad
We use random crop, random flip, and MixUp ($\alpha=0.8$) data augmentations, and label smoothing of $0.3.$ We train all MLP models for 500 epochs with batch size 1024, and all ViT models for 200 epochs with batch size 256. At the end of training, the models are close to but not exactly at convergence because fitting the training set is challenging due to strong augmentations and label smoothing. We do not use early stopping as it is not necessary.

We use structure-aware
learning rates and initialization described in \Cref{sec:struct-aware}, with a cosine learning rate decay to $0$. We set the constant in $\Theta(\cdot)$ as $1$ for the initialization standard deviations, with the exception that the last linear layer inside every residual block of the MLP and ViT is zero-initialized, as mentioned in \Cref{sec:struct-aware}. For a structured layer, zero-initialization is only applied to its last dense component so its output is zero at initialization but all the parameters receive non-zero gradients after the first step. Following \cite{yang2021v}, we also zero-initialize the classification layer and the query projection $\bm{W}_Q$ in transformers. We found zero-initialization generally improves performance.

We use a base learning rate of $\eta_0 = 3e-3$ for a dense MLP at $d_0=64,$ and $\eta_0 = 1e-3$ for a dense ViT at $d_0=64.$ For MLPs, we scale the learning rate of the input layer by a factor of $0.1$ since the input image dimension is much larger than $d_0$. This small multiplier prevents the first layer feature updates from having much larger scales than the other layers \citep{yang2023spectral}, which we found improves performance.

\noindent \textbf{Structure-specific hyperparameters.} \quad
We provide hyperparameters such as ranks we use for each structure and any other design choices we make.
\begin{itemize}
    \item Low-rank: we set the ranks of low-rank matrices to $\sqrt{\min(\din, \dout)}$ for MLP and $0.1 \times \min(\din, \dout)$ for ViT. The first choice leads to $\order{d^{3/2}}$ scaling of compute and parameters, same as Kronecker, 2-core BTT, and 2-core TT, but the second choice works significantly better for ViTs. We round the rank to its nearest integer when necessary. We initialize $\bm{V} \in \R^{r \times d}$ of the low-rank layer as $V_{ij} \sim \mathcal{N}(0, \sqrt{1/\din}), $ rather than $V_{ij} \sim \mathcal{N}(0, \sqrt{1/(r\din)}).$ While the latter is required for having the desired spectral norm at initialization according to \Cref{sec:struct-aware}, when we choose a rank of $\sqrt{\min(\din, \dout)},$ it is not compatible with our zero-initialization scheme as it led to vanishing gradients for both $\bm{U}$ and $\bm{V}$ as the width gets large.
    \item Kronecker: for any dimension $d$ that is not a perfect square, we factorize it so that the factors are as close as possible. For example, for a $20 \times 30$ matrix, we use the factorization $\bm{L} \otimes \bm{R}$ where $\bm{L} \in \R^{4 \times 5}$ and $\bm{R} \in \R^{5 \times 6}.$
    \item TT: we use two cores with TT-rank of $16$ for MLPs and $8$ for ViTs. We deal with non-perfect-square dimensions same as in Kronecker.
    \item Monarch: unless otherwise specified, we use $\bm{L}$ and $\bm{R}$ with 4 blocks, following the ViT and GPT-2 experiments in \citet{dao2022monarch}.
    \item BTT: we use BTT with various ranks and deal with non-perfect-square dimensions same as in Kronecker.
\end{itemize}

\section{Results for BTT with $c > 2$} \label{app:more-cores}
\begin{figure}[!h]
\centering
    \subfloat[MLP CIFAR-10 Train]{
    \includegraphics[width=0.4\linewidth]{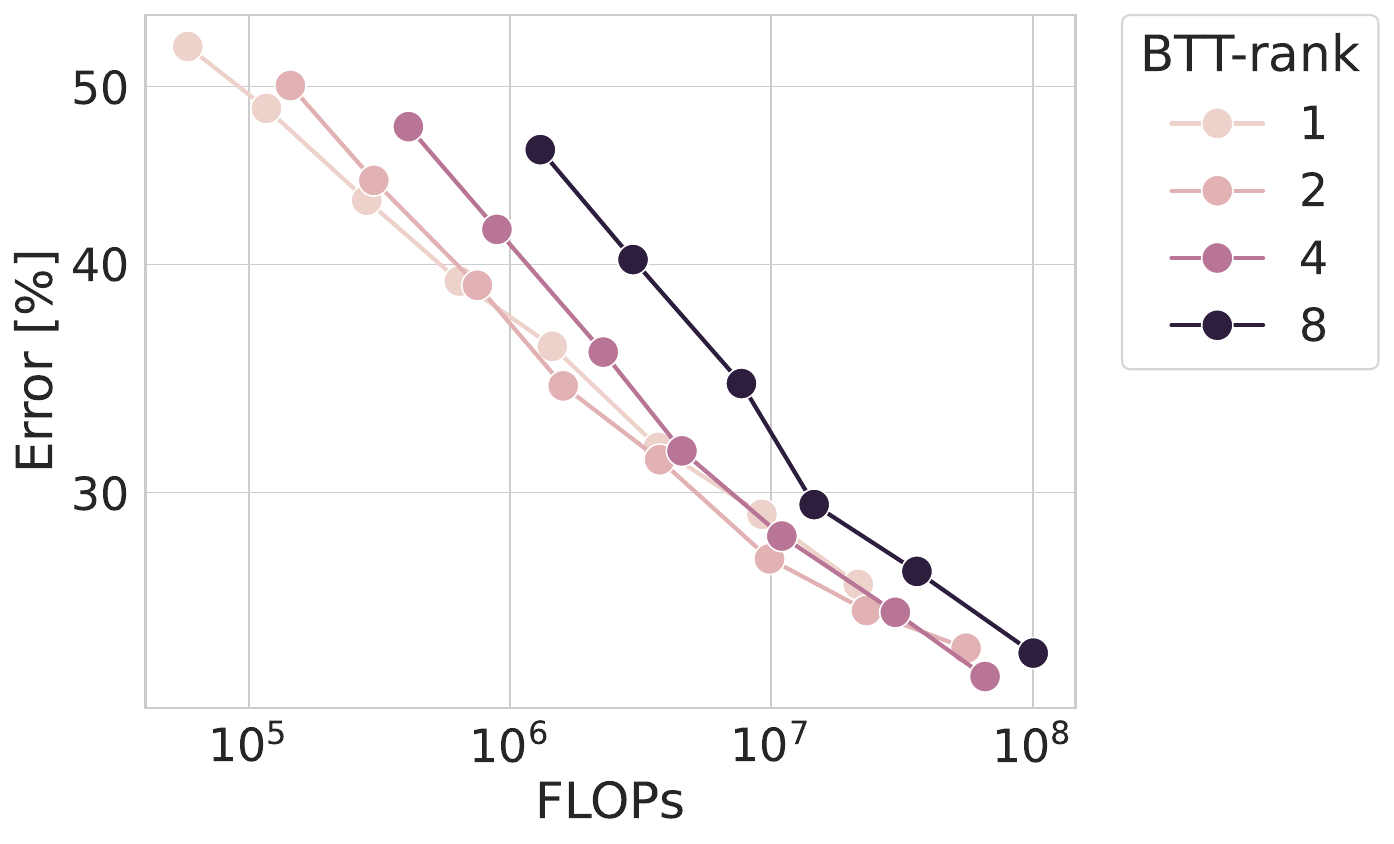}
    }
    \subfloat[ViT CIFAR-100 Train]{
    \includegraphics[width=0.4\linewidth]{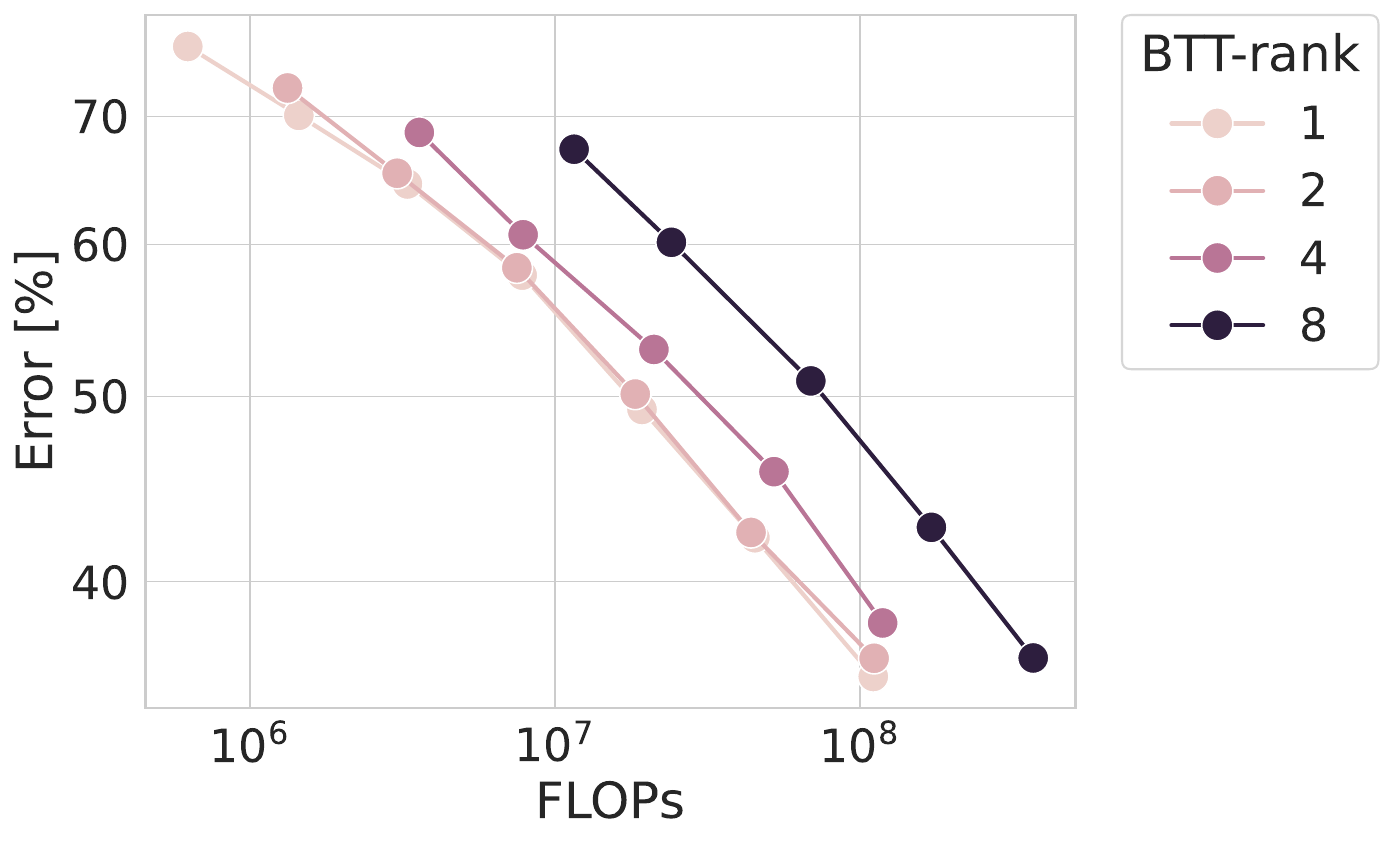}
    }
   \caption{ \textbf{Lower BTT-ranks have better compute-efficiency for BTT with $c=3$ cores.} Controlling for FLOPs, increasing the rank often degrades performance, though it reduces memory cost as the width is smaller.
   }
    \label{fig:3-core-rank}
    \vspace{-4mm}
\end{figure}

In \Cref{fig:compute_per_dim}, we showed scaling compute per dimension $\xi$ as $\xi=2d^{1/2}$ using BTT with $c=2$ and $r=1$ leads to better scaling laws than other choices of $r$ that increases $\xi$ to $2r^{1/2}.$ The gap between different choices of $r$ closes as the models are scaled up in width, e.g. $d \gg r.$
In \Cref{fig:3-core-rank}, we show a similar trend for $c=3,$ where higher values of $r$ perform worse when controlling for FLOPs, though the gap tends to vanish as the width is scaled up. Each connected line shows the performance of BTT with a fixed $r$ while $d$ is increased.

\begin{figure}[!h]
\centering
    \subfloat[MLP CIFAR-10 Train]{
    \includegraphics[width=0.4\linewidth]{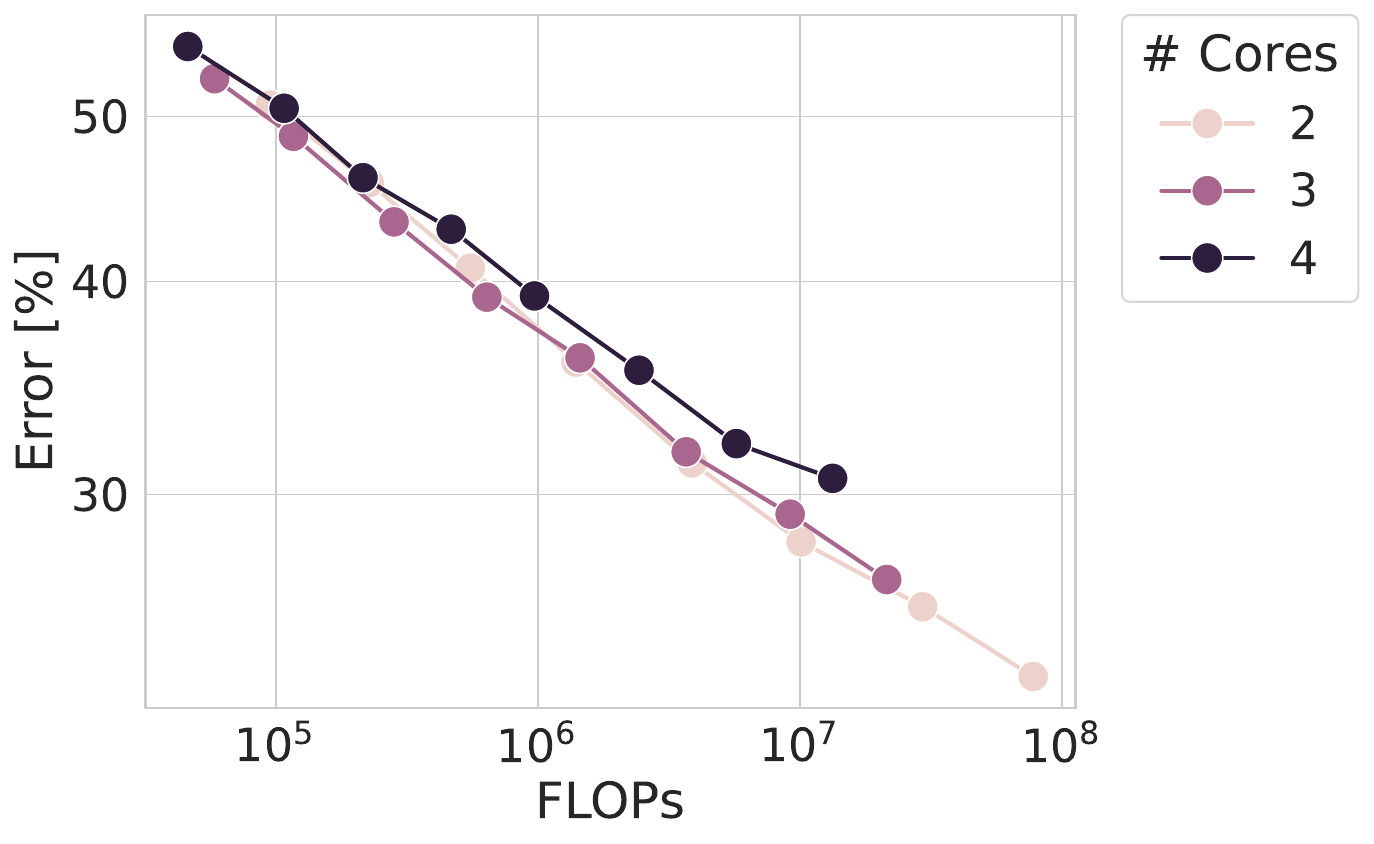}
    }
    \subfloat[ViT CIFAR-100 Train]{
    \includegraphics[width=0.4\linewidth]{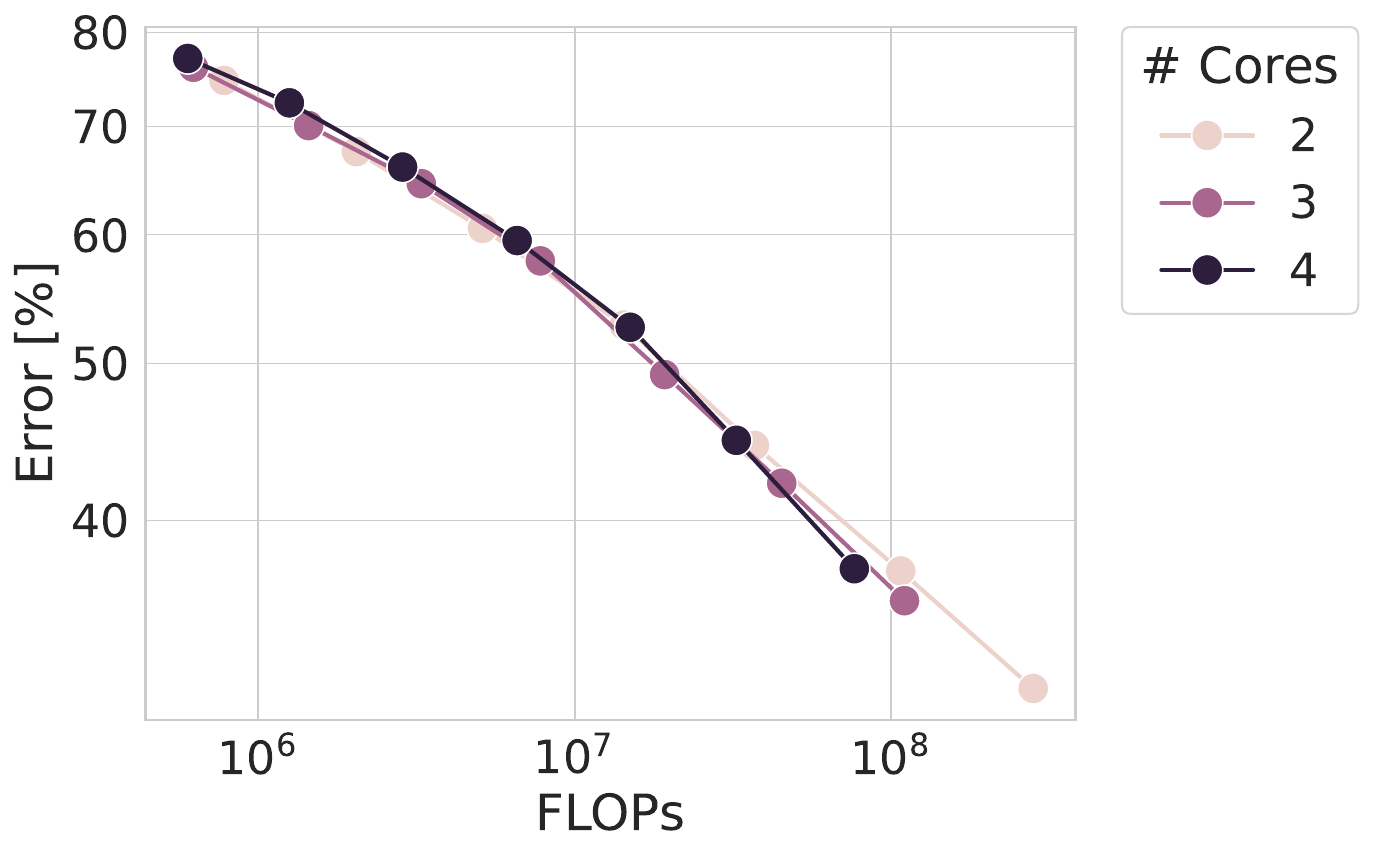}
    }
   \caption{ \textbf{BTT with $c=2$ cores achieves near-optimal compute-efficiency.} Controlling for FLOPs, increasing $c$ beyond 2 leads to no or negligible improvement in performance, while incurring higher memory costs as the models are wider.
   }
    \label{fig:more-cores}
    \vspace{-4mm}
\end{figure}

In \Cref{fig:more-cores}, we show the performance of BTT with $r=1$ and $c \in \{2, 3, 4\}.$ Further reducing the scaling of $\xi$ to $3d^{1/3}$ or $4d^{1/4}$ brings no or negligible improvement to performance when controlling for FLOPs.

In summary, choosing $c=2$ and $r=1$ leads to near-optimal performance for BTT on these tasks. In this case, BTT is equivalent to Monarch with $\sqrt{d}$ blocks.

\section{Transformer experiments} \label{app:trans}
We provide code for reproducing our experiments \href{https://github.com/shikaiqiu/compute-better-spent}{\underline{here}}.
\subsection{ViT on ImageNet}
We train with a global batch size of 3072 for 300 epochs with random crops, horizontal flip, random augmentations (\texttt{rand-m9-mstd0.5-inc1} from the \texttt{timm} library \cite{rw2019timm}), and Mixup of 0.2. The model has 12 transformer blocks, with width $d_\mathrm{model}$ ranging from $80$ to $384$ for dense. We use BTT with rank 1 or 2 and Monarch with 4 or 16 blocks. All but the classification head is replaced with structured matrices. We use the AdamW optimizer and set the base learning rate to $2e-3$ for the smallest dense model, which is transferred to other models via \mup and our structured-aware learning rate scaling. We apply a cosine learning rate decay to $0.$ The AdamW weight decay is set to $0.05$ for all models and is scaled automatically with width by being multiplied by the learning rate \citep{yang2021v}. The architecture is identical to the one in \Cref{app:arch}.

\subsection{GPT-2 on OpenWebText}
We train with a global batch size of 480 and a context length of 512 for 600,000 steps. We report the performance of the following models, all having 12 transformer blocks:
\begin{itemize}
    \item Structure $=$ Dense, $d_\mathrm{model}=384,$ $n_\mathrm{head} = 6$, $d_\mathrm{head} = 64$
    \item Structure $=$ Dense, $d_\mathrm{model}=512,$ $n_\mathrm{head} = 12$, $d_\mathrm{head} = 64$
    \item Structure $=$ Dense, $d_\mathrm{model}=768,$ $n_\mathrm{head} = 12$, $d_\mathrm{head} = 64$ (GPT-2 Small \citep{radford2019gpt2})
    \item Structure $=$ BTT ($r=4$), $d_\mathrm{model}=1024,$ $n_\mathrm{head} = 6$, $d_\mathrm{head} = 64$
    \item Structure $=$ BTT ($r=4$), $d_\mathrm{model}=1536,$ $n_\mathrm{head} = 6$, $d_\mathrm{head} = 64$
    \item Structure $=$ BTT ($r=4$), $d_\mathrm{model}=2048,$ $n_\mathrm{head} = 6$, $d_\mathrm{head} = 64$
    \item Structure $=$ BTT ($r=4$), $d_\mathrm{model}=2560,$ $n_\mathrm{head} = 12$, $d_\mathrm{head} = 64$
\end{itemize}

We use BTT with rank 4 in every linear layer, including the language modeling head. We set $n_\mathrm{head}$ to be smaller than the usual $d_\mathrm{model} / d_\mathrm{head}$ for the BTT models since otherwise we would spend too much compute in the attention layers relative to the FFN layers.  We use the Adam optimizer and set the base learning rate to $6e-4$ for the dense model at $d_\mathrm{model}=768$, which is transferred to other models via \mup and our structured-aware learning rate scaling.

\section{Structure-Aware Learning Rate for Other Optimizers} \label{app:other_opt}
The structure-aware learning rate scaling described in \Cref{sec:sucessful} applies to Adam or AdamW. However, we can derive appropriate scaling rules for other optimizers such as SGD. In Section 3.3, we obtain our structure-aware learning rate scaling rule in three steps: 1) decompose the matrix-vector multiplication (MVM) of a structured matrix $\bm{W} \in \R^{\dout \times \din}$ as a sequence of batched MVMs involving only dense matrices $\{\bm{G}_i\}_{i=1}^{k}$, 2) identify the input and output dimensions $d^i_\mathrm{in}$ and $d^i_\mathrm{out}$  of these dense matrices, 3) apply \mup to each of these dense matrices to scale their learning rates based on $d^i_\mathrm{in}$ and $d^i_\mathrm{out}$. Steps 1 and 2 are optimizer-agnostic. While step 3 is optimizer-dependent, it only requires knowing how to set \mup learning rates for regular dense matrices, which has been analyzed in prior works for various optimizers, including SGD, Adam, and SignSGD \citep{yang2023iv, yang2023spectral}. For example, instead of having the learning rate $\eta_i$ of $\bm{G}_i$ be $\Theta(1/d^i_\mathrm{in}),$ which is correct for Adam, SGD would require $\eta_i = \Theta(d^i_\mathrm{out} / d^i_\mathrm{in})$ \citep{yang2023spectral}. Therefore, the structure-aware learning rate multiplier relative to a dense $\bm{W}$ should now be $\kappa_i = \Theta\qty(\frac{d^i_\mathrm{out} / d^i_\mathrm{in}}{\dout / \din})$ instead of $\Theta(\din/d^i_\mathrm{in}),$ which is correct for Adam.

\section{Limitations and Future Work}
We provide a summary of the limitations of this work, and exciting directions for future work:
\begin{itemize}
\item Due to affordability constraints, we conducted our evaluation primarily with relatively small-scale models and datasets. Extending our evaluation to much larger-scale models and datasets is an important future direction.
\item The scaling laws we study differ from the compute-optimal scaling laws more relevant for large-scale training, which require optimally trading off between training larger models and training for more iterations. We only varied model size while keeping training iterations constant. Similarly, we did not optimize between scaling width v.s. depth, which allowed us to conveniently transfer learning rate through \mup\footnote{See \citet{yang2023tensor} for a depth extension of \mup and why it doesn't work for transformers in principle.}.
\item Our comparisons are based on FLOPs rather than runtimes. While the structures we consider have asymptotically the same MVM runtimes as dense matrices per FLOP (\Cref{app:runtime}), they introduce non-trivial runtime overhead for small matrix sizes, e.g. $\order{10^3}$. Developing highly optimized implementations will be important to realize the benefits of structured matrices in practice.
\item Despite our efforts to avoid over-fitting to image data (shuffling pixels for the MLP experiment),  our findings that structured matrices can significantly outperform dense matrices may still be highly dataset-dependent, as BTT offers a less significant improvement in language modeling compared to in image classification.
\item Our findings are empirical. Theoretically understanding when and why structured matrices can have better scaling laws than dense matrices, depending on model and data characteristics, will enable a prescriptive selection of structure rather than via trial and error alone.
\end{itemize}

\end{document}